\documentclass{article}

\usepackage[preprint]{neurips_2025}

\usepackage[utf8]{inputenc} 
\usepackage[T1]{fontenc}    
\usepackage{hyperref}       
\usepackage{url}            
\usepackage{booktabs}       
\usepackage{amsfonts}       
\usepackage{nicefrac}       
\usepackage{microtype}      
\usepackage{xcolor}         

\usepackage{subcaption}
\usepackage{multirow}%
\usepackage{graphicx}
\usepackage{tabularx}
\usepackage{longtable}

\graphicspath{ {./Figures/} }

\title{Examining Reject Relations in Stimulus Equivalence Simulations}

\author{%
Alexis Carrillo \\
Dep de Ps Clín, Psbiol y Metod \\
Universidad de la Laguna \\
\texttt{alu0101534640@ull.edu.es} \\
\And
Asieh Abolpour Mofrad \\
Department of Informatics \\
University of Bergen \\
\texttt{Asieh.Mofrad@uib.no} \\
\AND
Anis Yazidi \\
Department of Computer Science \\
Oslo Metropolitan University \\
\texttt{anisy@oslomet.no} \\
\And
Moises Betancort \\
Dep de Ps Clín, Psbiol y Metod \\
Universidad de la Laguna \\
\texttt{moibemo@ull.edu.es} \\
}

\begin{document}

\maketitle

\begin{abstract}
    Simulations offer a valuable tool for exploring stimulus equivalence (SE), yet the potential of reject relations to disrupt the assessment of equivalence class formation is contentious. This study investigates the role of reject relations in the acquisition of stimulus equivalence using computational models. We examined feedforward neural networks (FFNs), bidirectional encoder representations from transformers (BERT), and generative pre-trained transformers (GPT) across 18 conditions in matching-to-sample (MTS) simulations. Conditions varied in training structure (linear series, one-to-many, and many-to-one), relation type (select-only, reject-only, and select-reject), and negative comparison selection (standard and biased). A probabilistic agent served as a benchmark, embodying purely associative learning. The primary goal was to determine whether artificial neural networks could demonstrate equivalence class formation or whether their performance reflected associative learning. Results showed that reject relations influenced agent performance. While some agents achieved high accuracy on equivalence tests, particularly with reject relations and biased negative comparisons, this performance was comparable to the probabilistic agent. These findings suggest that artificial neural networks, including transformer models, may rely on associative strategies rather than SE. This underscores the need for careful consideration of reject relations and more stringent criteria in computational models of equivalence.
\end{abstract}

\begin{center}
\end{center}

\section{Introduction}\label{Introduction}

Simulations are valuable tools for analyzing the fundamental processes of Stimulus Equivalence (SE) and for generating novel hypotheses suitable for experimental analysis \citep{Tovar2023}. These simulations enable researchers to systematically control and manipulate virtual environments and computational agents to investigate critical variables influencing equivalence class formation, a complex phenomenon often difficult to analyze within traditional laboratory settings \citep{Lyddy2007, Mofrad2020, CarrilloBetancort2023}. This capacity for controlled manipulation can enhance our functional analysis of equivalence class formation and may lead to the development of more effective behavior-analytic interventions for educational and clinical applications \citep{tovar_westerman_2017}.

SE is a behavioral phenomenon characterized by the emergence of untrained stimulus relations. Specifically, when an organism is trained on a limited set of conditional discriminations, they may exhibit responding to a larger set of relations, including those not directly trained \citep{SidmanTailby1982, Sidman1994, Sidman2000}. In SE, organisms respond interchangeably to arbitrary and dissimilar stimuli as if those stimuli shared equivalent functions. The formation of equivalence classes enables the transfer of stimulus functions, such that a change in function for one member of an equivalence class will affect responding to all other members within that class \citep{GreenSaunders1998}.

Arbitrary Matching-to-Sample (MTS) is a procedure used to train conditional discriminations and to assess emergent relations in SE \citep{Arntzen2012, SidmanTailby1982}. In an MTS procedure, a sample stimulus is presented, followed by two or more comparison stimuli. The participant is required to select the comparison stimulus that corresponds to the sample. Correct responses are typically reinforced; incorrect responses may result in punishment or the absence of feedback \citep{GreenSaunders1998, Sidman2009}.

To establish an equivalence class with $M$ members, $M-1$ conditional discriminations are trained as baseline relations between sample stimuli and comparison stimuli. Equivalence class formation is demonstrated when all class members function as sample stimuli, exerting stimulus control within three-term contingencies in which all class members serve as discriminative stimuli. This control is assessed through the demonstration of reflexivity, symmetry, and transitivity. Consistent exhibition of these properties across all class members constitutes robust evidence for the establishment of a functional equivalence class \citep{Arntzen2012, GreenSaunders1998, Sidman1994, Sidman2000}.

Simulating SE involves training machine learning algorithms to replicate behaviors observed in human SE experiments \citep{Tovar2023}. These simulations employ various learning paradigms, including supervised \citep{CarrilloBetancort2023}, unsupervised \citep{garcia2010}, and reinforcement-learning approaches \citep{Mofrad2020, Mofrad2021}. Artificial neural networks (ANN), particularly Feedforward Networks (FFNs), have been frequently used as architectures in these simulations \citep{BarnesHampson1993, TovarTorresChavez2012, VernucioDebert2016, tovar_westerman_2017, Ninness2018}, with more recent studies also utilizing transformer-based models \citep{CarrilloBetancort2024}. Key methodological considerations and influencing variables in this research area include Training Structures (TS), the role of select and reject relations, and the specific algorithms employed as agents.

\begin{figure}[ht!]
    \centering
    \caption{Basic training structures for a class of four members (A, B, C, D). Baseline relations are shown in black solid arrows. Emergent relations for testing are reflexivity (self loop) in dashed grey arrows; symmetry in black dashed arrows; and transitivity in black dotted arrows }
    \includegraphics[width=0.95\linewidth]{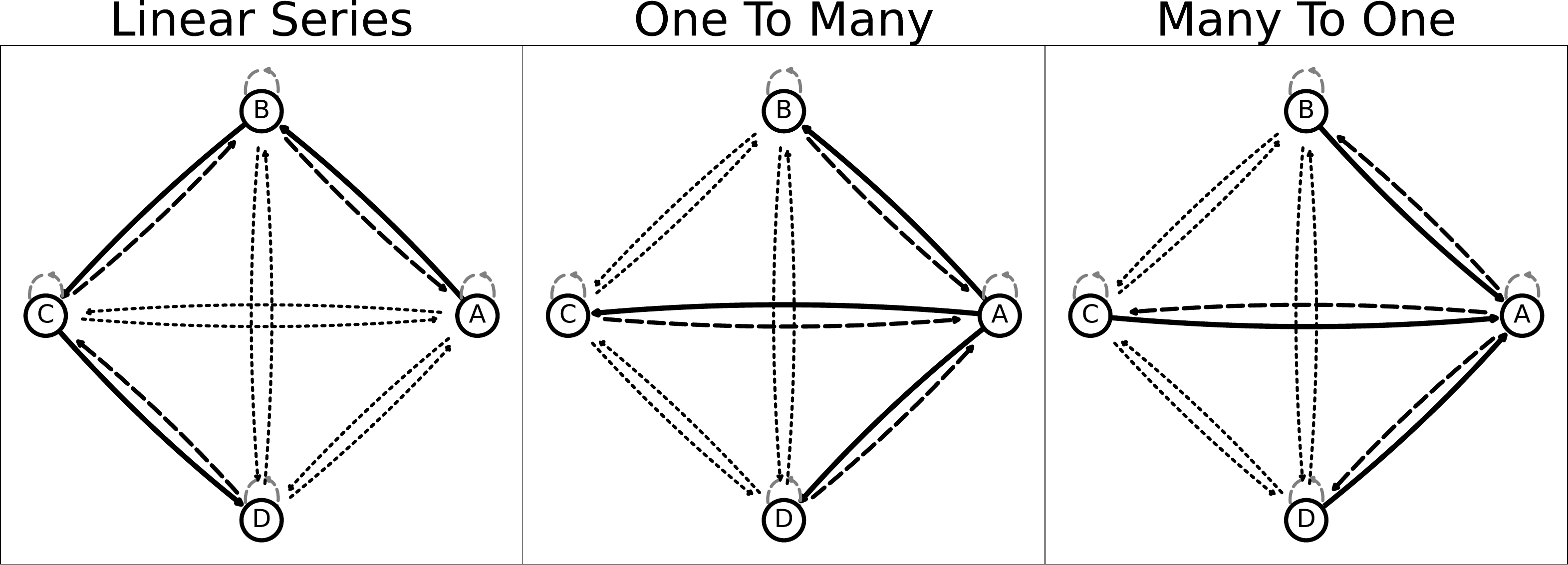}
    \label{fig:TS_introduction}
\end{figure}

TS refer to the specific arrangement in which conditional discriminations are presented during training and the trials used to assess equivalence class formation \citep{SaundersGreen1999}. In the formation of a four-member class (A, B, C, D), baseline pairs are selected as follows: A linear series (LS) presents stimuli in a sequential order (e.g., A-B, B-C, C-D). A many-to-one (MTO) structure presents multiple sample stimuli that are matched to a single comparison stimulus (e.g., B-A, C-A, D-A). A one-to-many (OTM) structure presents a single sample stimulus that is matched to multiple comparison stimuli (e.g., A-B, A-C, A-D) \citep{Arntzen2012, Ayres_PereiraVanessa2021}. Figure \ref{fig:TS_introduction} presents the baseline pairs (solid arrows) and their corresponding test pairs for the TS, as presented in \citet{CarrilloBetancort2024}.

\begin{figure}[ht!]
    \centering
    \caption{Example of a trial for the training of the A1-B1 pair with 3 comparison stimuli where B3 and B4 are incorrect comparison. Select relation is shown as a solid black arrow and reject relations as dashed grey arrows. Response indicate the action associated to the selection of a comparison. }
    \label{fig:trial_standard}
    \includegraphics[width=0.95\linewidth]{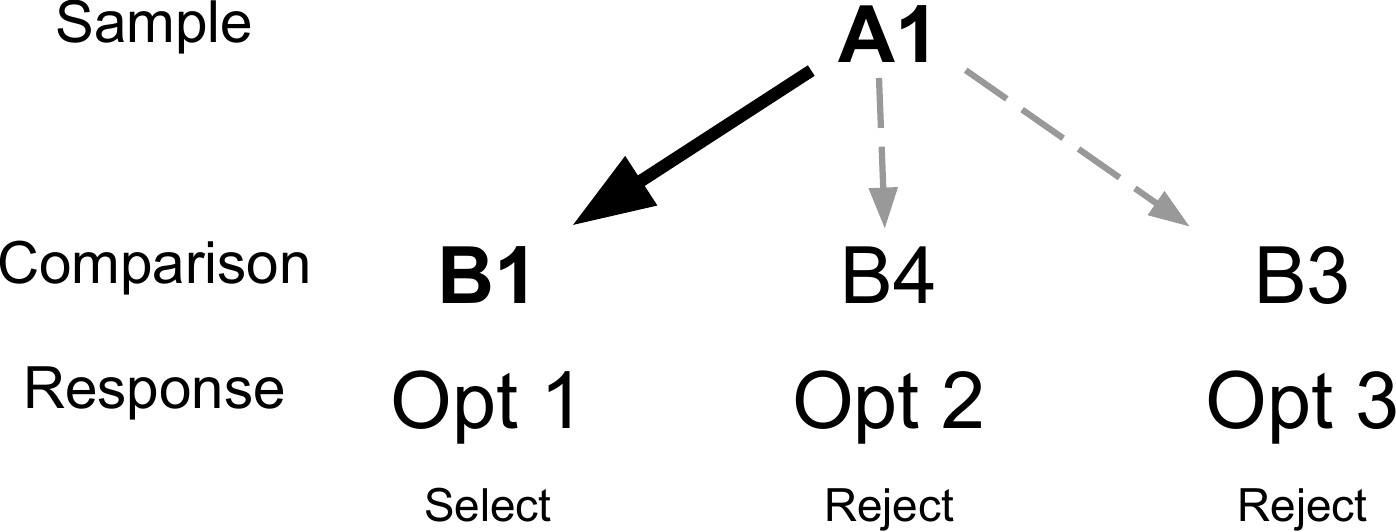}
\end{figure}

Correct selection of the comparison stimulus in MTS procedures can be driven by select relations (Sample/S+) and reject relations (Sample/S-) \citep{GreenSaunders1998}. Select and reject relations are two types of conditional discriminations that can be trained in baseline relations in an MTS task. A select relation refers to the learned association between a sample stimulus and the matching, or correct, comparison stimulus. This association is established through the selection of the matching stimulus with the consequence of a reward. An organism exhibiting a select relation consistently responds to the correct comparison stimulus when presented with a sample stimulus, indicating which comparison stimuli are appropriate matches for that sample. A reject relation is an association between a sample stimulus and a non-matching comparison stimulus, leading to no reward or punishment, thus signaling an incorrect choice. Participants tend to respond away from this comparison in the presence of the related sample \citep{CarriganSidman1992, JohnsonSidman1993}. Figure \ref{fig:trial_standard} presents the potential relations that can be established in a trial.

Control by the negative stimulus occurs when the participant's selection of the correct comparison stimulus is based on the avoidance of the negative comparison stimulus. This type of control can affect the emergence of equivalence relations and can lead to different responses patterns compared to when the positive stimulus is the main determinant of choice \citep{CarriganSidman1992, JohnsonSidman1993, Plazas2024}. Although control by select relations is typically associated with the emergence of equivalence relations \citep{CarriganSidman1992, Boldrin2022, Plazas2024}, there is an ongoing debate about whether select relations alone are sufficient for class formation or if the inclusion of reject relations is necessary \citep{Plazas2016, Plazas2018}.

Investigating the role of reject relations in SE presents methodological challenges, theoretical debates, and complexities in disentangling the interaction of select and reject relations. In the standard MTS procedure, the inherent interdependence of correct selection and the implicit rejection of alternative comparisons confounds the isolation of their respective contributions to equivalence class formation \citep{Sidman1994, GreenSaunders1998}.

The status of reject relations as equivalence relations, exhibiting reflexive, symmetric, and transitive properties analogous to select relations, remains a subject of ongoing debate within stimulus equivalence research \citep{Plazas2024, Boldrin2022}. \citet{CarriganSidman1992} posited that both select and reject relations constitute equivalence relations, potentially leading to the formation of distinct equivalence classes. This perspective suggests that reflexivity, symmetry, and transitivity can be observed under conditions of reject control, albeit with potential differences in behavioral expression compared to select control. Under reject control, reflexivity tests may manifest as oddity matching rather than the identity matching typically observed with select control. The influence of reject control on transitivity tests is predicted to be contingent on the number of nodes in baseline training. For tests involving an odd number of nodes, reject control is hypothesized to produce results inverse to those obtained under select control. 

\citet{JohnsonSidman1993} demonstrated oddity-like responding on reflexivity tests and contrasting outcomes on one-node transitivity tests when reject control was probable. \citet{Perez_Tomanari_2015, Perez_Tomanari_2020} also reported that manipulations inducing reject control affected performance on reflexivity and transitivity tests. However, this view has been challenged by subsequent research suggesting that reject control may even be beneficial or necessary for equivalence class formation \citep{Plazas2016, Plazas2018, Plazas2021}.

Sidman's proposition that conditional discrimination training with common responses and reinforcers initially establishes a broad class, subsequently differentiated into smaller classes \citep{Sidman1994, AlonsoAlvarez2023}, presents a methodological challenge. Distinguishing whether empirical data reflect this initial broad class formation or merely the failure to acquire conditional discriminations is problematic. Participants frequently exhibit mixed control, wherein some baseline relations are governed by select control and others by reject control \citep{CarriganSidman1992, JohnsonSidman1993}. This mixed control can yield inconsistent responses on equivalence tests, thereby complicating the precise determination of each relation type's contribution. Furthermore, humans, particularly those with prior experience in sorting and categorization, may demonstrate a pre-existing bias toward attending to select relations over reject relations \citep{Boldrin2022, Plazas2016, Plazas2024}.

Computational models and simulations offer an opportunity to investigate the dynamics of select and reject relations; however, these models require careful validation against human data and further refinement to ensure their biological plausibility \citep{BarnesHampson1993, Tovar2023, Mofrad2020}. \citet{VernucioDebert2016} used a go/no-go procedure with compound stimuli, responding to within-class stimuli and withhold responses to between-class stimuli. This procedure could help to explore \citet{Plazas2018} suggestion that between-class reject relations help establish a clear contrast between different stimulus classes. Mofrad's simulations, built on the Projective Simulation framework, successfully modeled equivalence class formation replicating prominent human experiments \citep{Mofrad2020, Mofrad2021}.

These models inherently utilize both select and reject control; however, they do not isolate or manipulate reject control as an independent variable. Isolating select and reject control in experimental settings typically involves modifications to the standard MTS protocol. \citet{CarrilloBetancort2024} implemented a select-only condition using dummy stimuli, based on the proposition by \citet{CarriganSidman1992} that SE can be demonstrated without explicit training of reject relations.

\begin{figure}[ht!]
    \centering
    \caption{Example of a trial for the training of the A1-B1 pair with D4 and C3 as negative comparisons from other classes with different labels than B. }
    \label{fig:trial_biased}
    \includegraphics[width=0.95\linewidth]{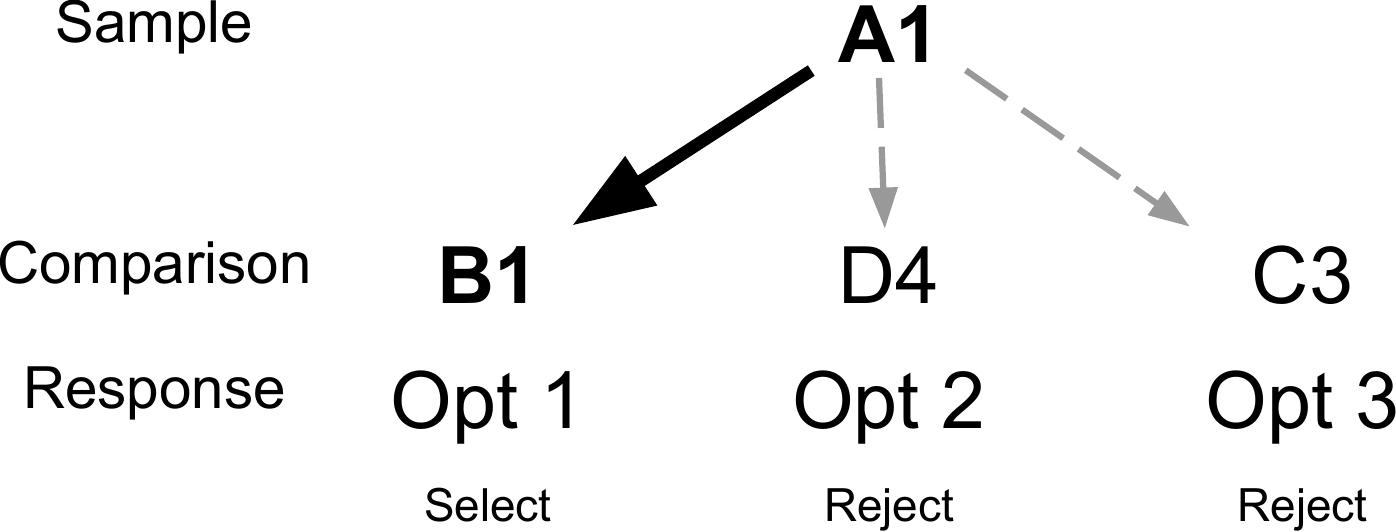}
\end{figure}

\citet{CarrilloBetancort2023} simulations using FFNs and transformer simulations employing the select-reject condition \citep{CarrilloBetancort2024} utilized all other class members from all labels as negative comparisons (see Figure \ref{fig:trial_biased}). This combinatorial expansion of trials, while potentially benefiting from data augmentation, deviates from the standard MTS procedure, which typically selects negative comparisons from other classes sharing the same label (e.g., selecting comparisons B2, B3, and B4 when training A1-B1). By incorporating all possible reject relations involving other class members, their approach introduced an imbalance wherein reject relations significantly outnumbered select relations. These additional reject relations, contingent on the chosen TS, could exert influence on equivalence test outcomes. Consequently, this departure from standard MTS procedures carries theoretical implications for the interaction between select and reject control in SE.

Consistent with the proposal of S+ and S- balance \citep{Plazas2016, Plazas2018}, an emphasis on S- relations may impede equivalence class formation. If both select and reject relations are conceptualized as equivalence relations \citep{Sidman1994, Sidman2000}, training all reject relations could be incompatible with classes established through select relations, potentially inducing ``oddity'' responding. From either theoretical standpoint, this methodological decision could confound simulation outcomes, obscure interpretations regarding class formation, and restrict comparability with standard MTS experiments.

TS, select-reject control type, and the number of negative comparisons encompasses different sets of reject pairs, potentially influences S- control by determining which reject pairs are trained. These factors directly impact the extent to which participants learn reject relations during training and subsequently affects their performance during testing. The selection of sample and comparison stimuli varies across TS (LS, OTM, MTO), altering the distribution of learned select and reject relations. In LS, both sample and correct comparison change; in MTO, the node stimulus serves as the correct comparison, concentrating select and reject relations on that stimulus. Conversely, OTM concentrates relations on the sample. A typical MTS procedure, involves both select and reject control, each potentially influencing different response patterns. Using all other class members as negative comparisons \citep{CarrilloBetancort2023, CarrilloBetancort2024}, includes $M \cdot (C-1)$ stimuli and overemphasizes reject relations, potentially leading to an over-reliance on S- control. This deviates from standard MTS, which uses a smaller number ($C-1$ stimuli) of negative comparisons.

The present study examines the hypothesis that reject relations can disrupt the assessment of equivalence class formation in simulations with computational agents. Although select relations are traditionally regarded as the foundation of SE, the role of reject relations remains contentious \citep{CarriganSidman1992, Plazas2018}. We propose that an excessive reliance on reject relations during training may enable agents to respond accurately on equivalence tests without demonstrating SE. Specifically, we hypothesize that agents may acquire reject relations and achieve correct responding on equivalence tests through simple associative learning, rather than derived relational responding indicative of class membership. This associative response pattern could produce accurate performance on symmetry and transitivity tests via the exclusion of incorrect comparisons, even in the absence of equivalence class formation. Consequently, this research seeks to systematically control the information provided to agents regarding the quantity of reject relations and S- control by manipulating negative comparisons and TS in SE simulations. By systematically varying these factors, we can evaluate their impact on agent performance and discern whether observed performance reflects equivalence class formation or a pattern of exclusion based on reject relations.

\section{Method}
\subsection{Design and variables} 
Simulations of MTS procedures were performed with variations across three factors: TS, select-reject control type, and number of negative comparisons. Three TS: LS, OTM, and MTO. Three relation types: Select-Reject, with both select (S+) and reject (S-) control is present, as typically observed in MTS procedures; Select-Only, where only S+ control was operative; and Reject-Only, where only S- control influenced the response. Two levels of negative comparisons: a standard MTS procedure in which the number of negative comparisons was limited to the members from the other classes with the same label, and a condition in which all other class members were presented as negative comparisons, as employed by \citet{CarrilloBetancort2023, CarrilloBetancort2024}, allowing for the training of a larger number of trials. Table \ref{tab:experimental_conditions} presents the 18 experimental conditions resulting from the combination of these three factors.

Each of the 18 experimental conditions represents a unique variation of reject relations. Three conditions replicate standard MTS procedures on each of the three TS (LS, OTM, and MTO) and serve as references for the other 15 experimental conditions, which are variations of this standard form. A total of 72 simulations were conducted, resulting from the execution of each of the 18 experimental conditions on four different agents.

All experimental conditions attempt to train the same four classes ($C$ = 1, 2, 3, 4), each consisting of six members ($M$ = A, B, C, D, E, F), resulting in 24 stimuli labeled and distributed to the classes as follows: (A1, B1, C1, D1, E1, F1), (A2, B2, C2, D2, E2, F2), (A3, B3, C3, D3, E3, F3), and (A4, B4, C4, D4, E4, F4).

\begin{figure*}[h]
    \centering
    \caption{Training structures for a 6 members class. Baseline pairs are shown in black solid arrows; reflexivity in self-loop dashed grey arrows; symmetry in black dashed arrows; and transitivity in black dotted arrows.}
    \label{fig:train_structures}
    \includegraphics[width=0.95\linewidth]{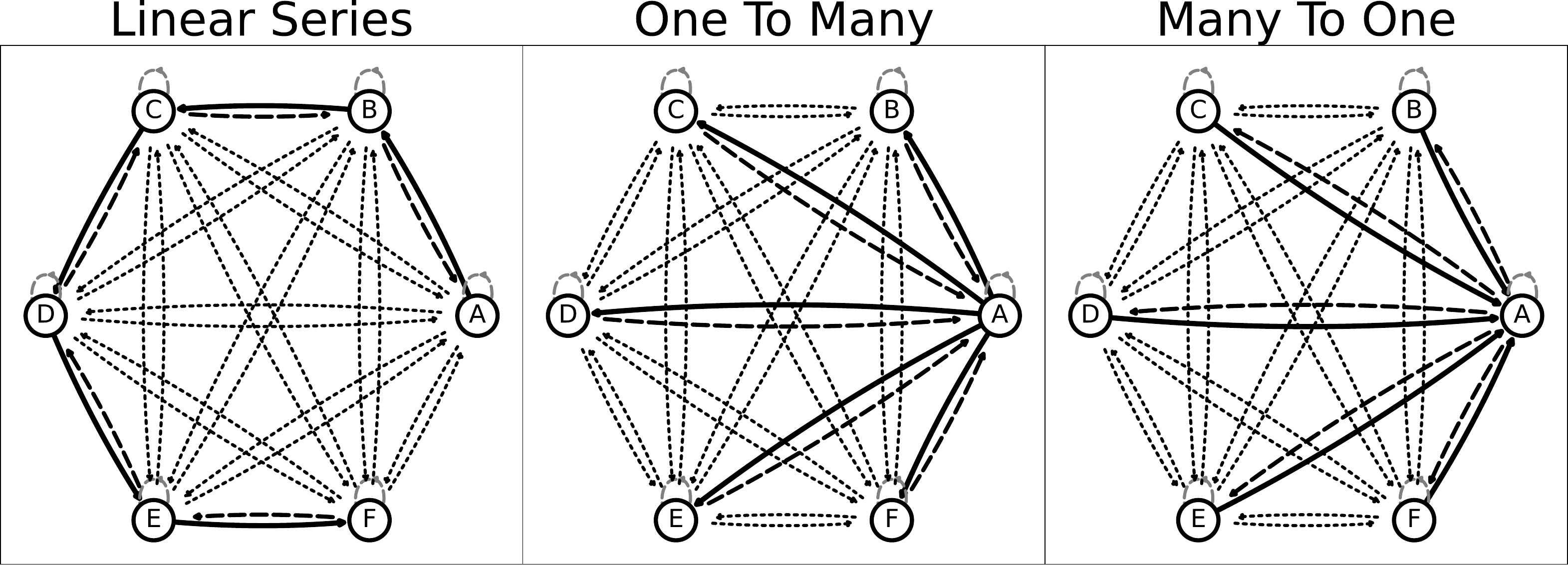}
\end{figure*}

The baseline relations in the experiment were defined based on the TS. For LS, baseline relations were sequential, consisting of pairs AB, BC, CD, DE, and EF. In OTM, stimulus A served as the sample for all other comparisons, resulting in pairs AB, AC, AD, AE, and AF. In MTO, stimulus A served as the comparison for all other samples, forming the pairs BA, CA, DA, EA, and FA. Reflexivity pairs were consistent across all three training structures, consisting of pairings of each stimulus with itself (e.g., AA, BB, CC, DD, EE, FF, GG). The pairs used to evaluate the properties of transitivity and symmetry were selected based on the specific TS, reflecting the different ways in which stimuli were related during the baseline training. Figure \ref{fig:train_structures} presents the pairs for baseline and evaluation for each of the TS.

In standard MTS procedures, negative comparisons typically consist of members from other equivalence classes. This establishes both select relations with same-class members and reject relations with members of other classes during training. This configuration is referred to as the Select-Reject condition in the simulations. To examine the influence of different types of relations, variations of the standard MTS procedure were developed. An additional set of ($ M \cdot C $) 24 dummy stimuli, labeled with the letter ``Z'', an underscore character, and numbers from 11 to 34 (Z\_11, Z\_12, Z\_13 ... Z\_34), were introduced.

In the Select-Only variation \citep{CarrilloBetancort2024}, negative comparisons during training were replaced with these dummy stimuli. This manipulation prevented the formation of reject relations with members of other classes. Consequently, during the evaluation phase, information about reject relations was unavailable, allowing us to assess equivalence class formation based solely on select relations.

\begin{figure}[ht!]
    \centering
    \caption{Trials exemplifying relation type variations. On select-only trials (a), Z\_23 and Z\_17 are dummy stimuli used as negative comparisons on baseline train. On reject-only trials (b), Z\_12 is a dummy stimuli used as positive comparison on baseline train.}
    \label{fig:trials_relation_types}
    \begin{subfigure}{0.95 \linewidth}
        \caption{Select only trial}
        \includegraphics[width=\linewidth]{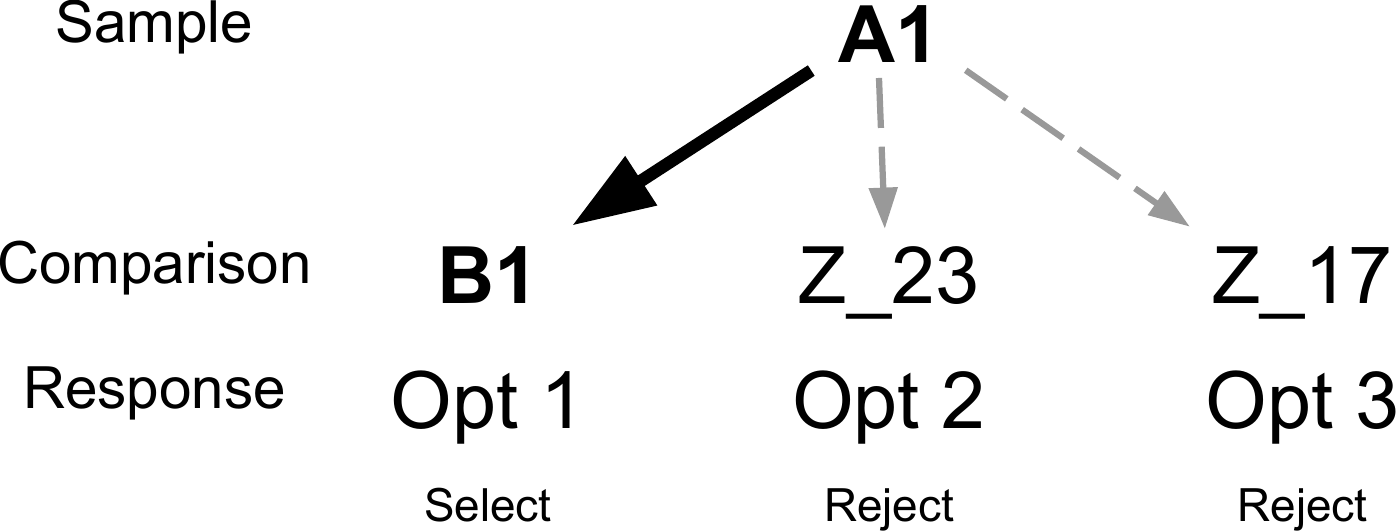}
        \label{fig:trial_select_only}
    \end{subfigure}
    \\[15pt] 
    \begin{subfigure}{0.95 \linewidth}
        \caption{Reject only trial}
        \includegraphics[width=\linewidth]{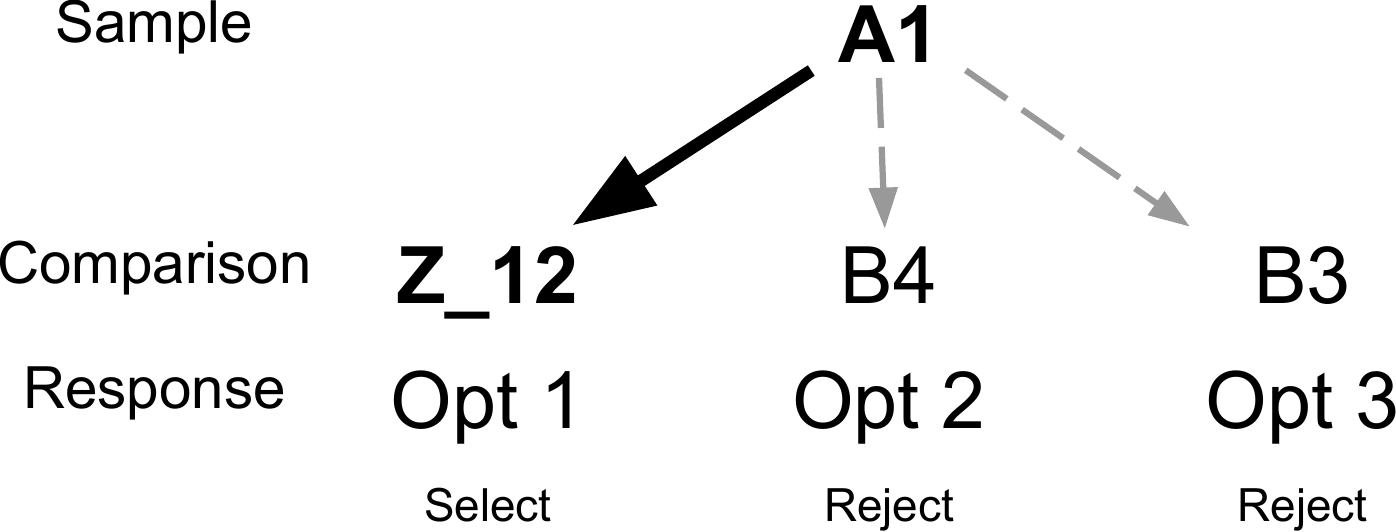}
        \label{fig:trial_reject_only}
    \end{subfigure}
\end{figure}

In the Reject-Only variation, positive comparisons were replaced with dummy stimuli during training. This manipulation prevented the formation of select relations within the same class. Consequently, during the evaluation phase, information about select relations was unavailable, allowing us to assess equivalence class formation primarily based on reject relations. Figure \ref{fig:trials_relation_types} presents trials for the MTS variations: select-only (\ref{fig:trial_select_only}) and reject-only (\ref{fig:trial_reject_only}).

In typical MTS procedures, negative comparisons are often selected from stimuli that share the same member label as the positive comparison but are members of different classes. According to this principle, in our simulations, for the baseline pair A1-B1, the negative comparisons are B2, B3, and B4. This approach is referred to as the ``Standard'' method of selecting negative comparisons in our paper.
We defined a ``Biased (S-)'' variation for selecting negative stimuli, where all other class members are used as negative comparisons~\citep{CarrilloBetancort2023, CarrilloBetancort2024}. This approach incorporates a larger number of reject relations into the training process compared to the standard selection method. For the A1-B1 pair, the negative comparisons include A2, B2, C2, D2, E2, F2, A3, B3, C3, D3, E3, F3, A4, B4, C4, D4, E4, and F4. See Appendix~\ref{Apx_relations} for a detailed comparison of the select and reject relations across the 18 experimental conditions listed in Table~\ref{tab:experimental_conditions}. Each cell in the table indicates a unique experimental condition, the name of which we will use to refer to that specific condition, with labels in parentheses referencing corresponding figures in the appendix that visually depict the trained relations for that condition.

\begin{table}[h]
\centering
\caption{Experimental conditions resulting from the combination of training structures (TS), relation types (Rel), and negative comparison selection (NCS) methods. Each cell lists the abbreviated name of the condition, with corresponding figure references (showing trained select and reject relations) in parentheses. See Appendix~\ref{Apx_relations} for full visualizations.}
\label{tab:experimental_conditions}
\begin{tabular}{@{}clll@{}}
\toprule
\multirow{2}{*}{\textbf{TS}} & \multicolumn{1}{c}{\multirow{2}{*}{\textbf{Rel}}} & \multicolumn{2}{c}{\textbf{NCS}} \\ \cmidrule(l){3-4} 
 & \multicolumn{1}{c}{} & \textbf{Standard} & \textbf{Biased (S-)} \\ \midrule
\multirow{3}{*}{\textbf{LS}} & \textbf{Select-Reject} & LS (\ref{fig:LS_standard}) & LS biased (\ref{fig:LS_biased}) \\
 & \textbf{Select-Only} & LS Select (\ref{fig:LS_select}) & LS Select biased (\ref{fig:LS_select_biased}) \\
 & \textbf{Reject-Only} & LS Reject (\ref{fig:LS_reject}) & LS Reject biased (\ref{fig:LS_reject_biased}) \\
\multirow{3}{*}{\textbf{OTM}} & \textbf{Select-Reject} & OTM (\ref{fig:OTM_standard}) & OTM biased (\ref{fig:OTM_biased}) \\
 & \textbf{Select-Only} & OTM Select (\ref{fig:OTM_select}) & OTM Select biased (\ref{fig:OTM_select_biased}) \\
 & \textbf{Reject-Only} & OTM Reject (\ref{fig:OTM_reject}) & OTM Reject biased (\ref{fig:OTM_reject_biased}) \\
\multirow{3}{*}{\textbf{MTO}} & \textbf{Select-Reject} & MTO (\ref{fig:MTO_standard}) & MTO biased (\ref{fig:MTO_biased}) \\
 & \textbf{Select-Only} & MTO Select (\ref{fig:MTO_select}) & MTO Select biased (\ref{fig:MTO_select_biased}) \\
 & \textbf{Reject-Only} & MTO Reject (\ref{fig:MTO_reject}) & MTO Reject biased (\ref{fig:MTO_reject_biased}) \\ \bottomrule
\end{tabular}
\end{table}

\subsection{Computational agents} 

A single-layer FFN with 50,000 hidden units received the encoded trial as input and produced an output vector with three units, each representing a response option. The network was trained using the Adam optimizer with a learning rate of 0.001. Training was allowed for a maximum of 50,000 epochs, but an early stopping mechanism based on the root mean squared error (RMSE) was used. Training was immediately stopped if the RMSE on the training set dropped below 0.001. Hyperparameter selection was based on prior testing and experimentation. This process helped identify a set of parameters that produced stable responses and achieved performance above  the mastery criterion on the baseline trials.

A Generative Pre-trained Transformer (GPT) model was employed, based on the architecture described in \citet{Karpathy2023NanoGPT} and implemented using the NanoGPT codebase. GPT, a transformer-based language model designed for autoregressive tasks, utilizes decoder blocks \citep{radford2018gpt1, radford2019gpt2}.

A Bidirectional Encoder Representations from Transformers (BERT) model was implemented using the NanoGPT codebase \citep{Karpathy2023NanoGPT}. A modification to the code, specifically the removal of the mask in the self-attention heads, converts the decoder blocks into encoder blocks. BERT, designed for bidirectional context understanding, utilizes encoder blocks \citep{devlin2019bert}. 

The following hyperparameters were used for training both BERT and GPT models: batch size of 64, block size of 4, maximum iterations of 10,000, evaluation interval of 500 iterations, learning rate of $3e-4$, evaluation iterations of 200, embedding size of 384, number of attention heads of 6, number of blocks of 6, and dropout probability of 0.2. We utilized the default hyperparameters defined by \citet{Karpathy2023NanoGPT} in his NanoGPT codebase. As Karpathy notes, these default parameters were themselves the product of experimentation and represent a reasonable starting point for training transformer models. While not explicitly tuned for the specific task in the present study, these established values provided a practical and principled approach to hyperparameter selection for BERT and GPT, allowing us to focus on the impact of our experimental manipulations rather than engaging in extensive hyperparameter optimization.


The probabilistic agent served as a benchmark, embodying a purely associative learning response pattern based on select and reject relations. It learns by associating sample-comparison pairs with positive or negative/no-reward outcomes, memorizing which comparisons to select (p = 1) and which to avoid (p = 0) for a given sample stimulus. It stores a matrix of probabilities, $P_{r+}( Comparison | Sample)$, for all possible stimulus pairs. This matrix is initialized with all values set to 0.5, signifying that the probability of reward is uncertain. A random number generated from a normal distribution (M = 0, SD = 0.01) is added to each initial probability to randomly break potential initial ties between equally probable pairs. During training, the probabilities are updated: for reinforced pairs, the probability is set to 1; for non-reinforced pairs, the probability is set to 0. On evaluation, the algorithm selects the stimulus pair with the highest probability of reward.

Serving as a control, the probabilistic agent provides a baseline that helps assess whether the other agents are responding with a response pattern based on associative learning of select and reject relations. We reasoned that any agent demonstrating equivalence class formation should pass equivalence tests across the experimental conditions and outperform the probabilistic agent. If an agent's performance is comparable to that of the probabilistic agent, particularly in conditions emphasizing reject relations, it suggests reliance on a similar associative response pattern.

\subsection{Apparatus} 
The experiments were conducted on a Lenovo LOQ 15" laptop (Intel Core i5-12450H, NVIDIA GeForce RTX 4060 8GB GDDR6). The code was implemented in Python 3 using the PyTorch deep learning framework \citep{paszke2019pytorch}. The code used to conduct the simulations reported in this study is publicly available at the following GitHub repository: \url{https://github.com/Yagwar/stim_eq/tree/master/reject_simulations}.

\subsection{Procedure} 
\subsubsection{MTS procedure} 

The MTS procedure involved the simultaneous presentation of one sample stimulus and three comparison stimuli. Across all 18 simulations, stimulus pairs were initially assigned to training and evaluation groups, consistent with the specified TS. Subsequently, trials were generated for each condition based on select-reject control type and negative comparison selection method. During the training phase, agents were presented with trials containing baseline stimulus pairs, receiving feedback after each trial. Agents advanced to the evaluation phase only upon achieving a minimum of 90\% correct responses. In the evaluation phase, trials assessing reflexivity, symmetry, and transitivity were presented without feedback, and agent responses were recorded. A correct response was defined as the selection of the appropriate comparison stimulus. Evaluation trials followed the standard MTS protocol (select-reject condition and standard negative comparison selection) to minimize error variance across MTS protocol variations and maintain comparability with existing SE research.

The primary performance measure was the correct response rate across all evaluation trials. A mastery criterion of 90\% correct responses was required to pass the test. Recognizing that computational models may not always perfectly replicate human performance, a less stringent criterion of 70\% correct responses was also employed to identify conditions where agents approached mastery, even if they did not meet the 90\% threshold. This 70\% criterion served as a marker of  ``near-mastery'' performance, enabling a more nuanced analysis of agent responding \citep{CarrilloBetancort2024}.

\subsubsection{Agents train and test} 
Class members and dummy stimuli were processed using a one-hot encoding scheme as binary vector for the FFN. Inputs consisted of the concatenated encodings of each stimulus within the trial. Output layer of three units, each representing a response option, also encoded using one-hot encoding. During training, the FFN received the encoded trial as input, and the feedback signal consisted of the encoded target values representing the correct response. On the evaluation phase, input values were passed to the FFN and the unit with the maximum activation value from the output layer was considered as the comparison selected. During the evaluation phase, input values were passed to the FFN, and the unit with the maximum activation value in the output layer was considered the selected comparison.

Following \citet{CarrilloBetancort2024} procedure, for the Transformer-based models (BERT and GPT), trials were represented as sequences of tokens using a vocabulary consisting of class members, dummy stimuli, and response options (O\_1, O\_2, O\_3). Each trial was considered a text sequence for the model, comprising five tokens: the sample stimulus, the first comparison stimulus, the second comparison stimulus, the third comparison stimulus, and the selected response option. The sequence of the first four tokens served as the context window for the model's training, with the response option token treated as the target token. During training, BERT and GPT received the sequence of tokens corresponding to the trial as input, and the target value consisted of the token representing the correct response option. During the evaluation phase, the tokens of the trial were presented as the prompt to the model. The first token generated by the agent was considered the selected comparison.

For the probabilistic agent, during training, the stimulus of the trial was presented as input, and the target value corresponded to the value of reward for the selection of each comparison: 1 for the correct comparison and 0 for the others. During evaluation, each trial was passed to the agent, and it returned probability values for each comparison. The comparison with the maximum probability value was selected.

\section{Results}

The present study employed a single simulation run for each agent within each of the 18 experimental conditions. This design choice prioritized a comparative analysis of the different experimental conditions and the performance of the distinct agent algorithms. Because each agent's responding is governed by a unique algorithm, an individual-level analysis was deemed most appropriate. Our analysis focused on the impact of variations in select and reject relations on the response patterns observed in the agents. A comparison against the performance of the probabilistic agent was conducted to probe whether the agents rely on an associative response pattern.

Appendix \ref{Apx_results}, Table \ref{tab:performance_complete}, presents the performance metrics for all 72 simulations, including baseline, reflexivity, symmetry, and transitivity tests. Values in bold indicate performance at or above the mastery criterion of 0.9. Across all training structures, agents consistently achieved mastery during baseline training. However, successful performance on reflexivity, symmetry, and transitivity tests was limited. Table \ref{tab:summary_performance} provides a qualitative description of agent performance across the experimental conditions outlined in Table \ref{tab:experimental_conditions}.

\begin{table*}[!t]
    \centering
    \caption{Summary of Test Performance Across Experimental Conditions} 
    \label{tab:summary_performance}
    \begin{tabularx}{\linewidth}{@{}l>{\centering\arraybackslash}p{1.5cm}X@{}} 
        \toprule
        \textbf{Condition} & \textbf{Sim.} & \textbf{Performance on Tests} \\
        \midrule
        LS (\ref{fig:LS_standard}) & 1-4 & Low across all agents (below 0.62) \\
        LS biased (\ref{fig:LS_biased}) & 5-8 & BERT mastery (0.99, 0.97, 0.95), FFN, GPT, Probabilistic high but below mastery \\
        LS Select (\ref{fig:LS_select}) & 9-12 & Consistently low (below 0.36) \\
        LS Select biased (\ref{fig:LS_select_biased}) & 13-16 & Low \\
        LS Reject (\ref{fig:LS_reject}) & 17-20 & Generally low \\
        LS Reject biased (\ref{fig:LS_reject_biased}) & 21-24 & Generally low, except Probabilistic agent (above 0.80) \\
        MTO (\ref{fig:MTO_standard}) & 25-28 & Generally low, except GPT symmetry (0.82) \\
        MTO biased (\ref{fig:MTO_biased}) & 29-32 & FFN and Probabilistic mastery on transitivity (0.93, 1.00), Probabilistic mastery on reflexivity (0.93), other scores low \\
        MTO Select (\ref{fig:MTO_select}) & 33-36 & Low \\
        MTO Select biased (\ref{fig:MTO_select_biased}) & 37-40 & Low \\
        MTO Reject (\ref{fig:MTO_reject}) & 41-44 & Low \\
        MTO Reject biased (\ref{fig:MTO_reject_biased}) & 45-48 & Probabilistic mastery on reflexivity and transitivity (0.89, 1.00) \\
        OTM (\ref{fig:OTM_standard}) & 49-52 & Low, except GPT symmetry (0.64) \\
        OTM biased (\ref{fig:OTM_biased}) & 53-56 & Low \\
        OTM Select (\ref{fig:OTM_select}) & 57-60 & Low \\
        OTM Select biased (\ref{fig:OTM_select_biased}) & 61-64 & Low \\
        OTM Reject (\ref{fig:OTM_reject}) & 65-68 & Low \\
        OTM Reject biased (\ref{fig:OTM_reject_biased}) & 69-72 & Low \\
        \bottomrule
    \end{tabularx}
\end{table*}

Only six out of the 72 simulations exhibited performance above the mastery criterion on at least one test. The distribution of these simulations across TS was as follows: Simulations were present in LS, MTO had the remaining three, and none of the simulations were found in OTM. By relation type, five simulations occurred in the Select-Reject condition, one in the Reject-Only condition, and none in the Select-Only condition. All six simulations were conducted with the biased selection of negative comparisons. While performances above 0.7 did not meet the stringent mastery criterion, they can still provide relevant information when analyzing results from computational agents \citep{CarrilloBetancort2024}. Table \ref{tab:relevant_tests} presents the nine simulations in which at least one reflexivity, symmetry, or transitivity test had a correct selection rate above 0.7.

\begin{table}[h]
    \centering
    \caption{Selected simulations with a correct selection rate (CSR) above 0.7 in any of the reflexivity (Ref), symmetry (Symm), or transitivity (Tran) tests. Simulations are identified by their number (N), Agent and experimental condition. Bold values indicate performance at or above the mastery criterion of 0.9.}
    \label{tab:relevant_tests}
    \begin{tabular}{@{}rllrrr@{}}
        \toprule
        \multicolumn{1}{l}{\multirow{2}{*}{\textbf{N}}} & \multirow{2}{*}{\textbf{Agent}} & \multirow{2}{*}{\textbf{Condition}} & \multicolumn{3}{l}{\textbf{Test}} \\ \cmidrule(l){4-6} 
        \multicolumn{1}{l}{} &  &  & \multicolumn{1}{l}{\textbf{Ref}} & \multicolumn{1}{l}{\textbf{Symm}} & \multicolumn{1}{l}{\textbf{Tran}} \\ \midrule
        5 & BERT & LS biased & \textbf{0.99} & \textbf{0.97} & \textbf{0.95} \\
        6 & FFN & LS biased & 0.89 & 0.87 & 0.87 \\
        7 & GPT & LS biased & 0.72 & \textbf{0.96} & 0.84 \\
        8 & Probabilistic & LS biased & 0.88 & \textbf{0.95} & 0.86 \\
        24 & Probabilistic & LS reject biased & 0.85 & 0.88 & 0.85 \\
        27 & GPT & MTO & 0.04 & 0.82 & 0.04 \\
        30 & FFN & MTO biased & 0.84 & 0.34 & \textbf{0.93} \\
        32 & Probabilistic & MTO biased & \textbf{0.93} & 0.23 & \textbf{1.00} \\
        48 & Probabilistic & MTO reject biased & 0.89 & 0.45 & \textbf{1,00} \\ \bottomrule
    \end{tabular}
\end{table}

The probabilistic agent, serving as a benchmark for associative learning, exhibited varying performance across conditions (see appendix \ref{Apx_results}). On LS biased, the probabilistic agent achieved 88\% correct selection on reflexivity tests, 95\% on symmetry tests, and 86\% on transitivity tests. With LS reject biased condition, performance remained above 80\% for all tests, reaching 85\% on reflexivity, 88\% on symmetry, and 85\% on transitivity. However, performance in the LS select biased condition dropped substantially, with accuracy below 40\% for all three tests. In the standard negative comparison conditions within the LS structure, performance was consistently low across all relation types, with accuracy ranging from 25\% to 38\%. Within the MTO training structure, the probabilistic agent's performance was more variable. With select-reject relations and biased negative comparisons, it achieved 93\% on reflexivity and perfect performance (100\%) on transitivity, but only 23\% on symmetry. In the reject-only condition with biased negative comparisons, the probabilistic agent again showed high performance on reflexivity (89\%) and transitivity (100\%), but low performance on symmetry (45\%). The standard negative comparison conditions within the MTO structure yielded consistently low performance across all relation types. Across all simulations employing the one-to-many (OTM) training structure, the probabilistic agent’s performance was consistently below 50\% for all tests, regardless of relation type or negative comparison strategy.

The FFN agent, trained with the LS biased condition, demonstrated performance comparable to the probabilistic agent, achieving 89\% on reflexivity, 87\% on symmetry, and 87\% on transitivity. With the MTO structure and select-reject relations and biased negative comparisons, the FFN performed well on reflexivity (84\%) and transitivity (93\%), but not on symmetry (34\%). Similar to the probabilistic agent, the FFN agent generally did not demonstrate strong equivalence class formation in other conditions. Performance was low in the OTM structure regardless of relation type and negative comparison strategy, and scores were consistently below 50\%. Likewise, performance in the standard negative comparison conditions across both LS and MTO structures ranged from 30\% to 35\%.

BERT showed the strongest performance, but only under a specific condition. When trained with the LS biased condition, BERT achieved scores above the mastery criterion on all equivalence tests: 99\% on reflexivity, 97\% on symmetry, and 95\% on transitivity. In all other conditions, BERT’s performance was substantially lower, with scores generally below 40\%.

GPT agent's performance was also variable. In the LS biased condition, GPT achieved 76\% on reflexivity, 96\% on symmetry, and 84\% on transitivity. Notably, GPT was the only agent to achieve a score above 70\% in a standard negative comparison condition, reaching 82\% on symmetry in the MTO condition. However, GPT's performance was low on all other tests.

Across all simulations employing the OTM training structure, no agent achieved a performance score at or above the mastery criterion of 90\%, nor even the less stringent threshold of 70\%. This pattern was consistent across both standard and biased negative comparison selection methods, indicating uniformly low performance regardless of how reject relations were manipulated within this training structure. Furthermore, within the biased negative comparison condition, only the isolated instances of near-mastery performance described above for the probabilistic, FFN, and GPT agents were observed.

\section{Discussion}
This study examined whether a predominance of reject control during baseline training could result in accurate responding on equivalence tests in connectionist models in the absence of emergent equivalence classes. We manipulated the salience of S- relations during the conditional discrimination training phase to evaluate their influence on the response patterns of artificial neural network agents. Across simulations employing FFN, BERT, and GPT architectures, limited evidence for the formation of stimulus equivalence classes was observed. Our findings suggest that the manipulation of S- relations and varying sample stimuli during baseline training significantly influenced agent performance on subsequent equivalence tests. This suggesting the potential for rule-based responding based on exclusion rather than the emergent properties of equivalence. Three key findings support this interpretation. 

First, agents demonstrated consistent failures in simulations where S- information was limited or unavailable during training, as observed in the select-only condition and the standard selection of negative comparisons.

Second, successful performance on test trials was predominantly observed in the LS and MTO training structures. These structures, particularly when combined with select-reject conditions and biased negative comparisons, potentially allowed agents to learn more salient S+ and a greater number of S- associations through differential reinforcement histories \citep{Ayres_PereiraVanessa2021}. The biased negative comparison conditions, exposing agents to a higher frequency of non-reinforced comparisons, might have fostered a response pattern based on exclusion \citep{Plazas2021}. The relative absence of successful simulations in the OTM structure further supports this interpretation, as OTM, unlike LS and MTO, involves a training history where a single sample stimulus is consistently paired with multiple comparisons, potentially limiting the opportunity to learn specific S- relations in the same manner. 

Third, the performance of BERT, GPT, and FFN agents was comparable to that of a probabilistic benchmark agent. This similarity in performance between complex ANN architectures and a purely associative benchmark suggests that the ANNs may have also based their responses on learned associations between stimuli based on reinforcement contingencies, rather than on the emergent properties of equivalence relations \citep{SidmanTailby1982}. This outcome underscores that accurate baseline performance on conditional discriminations does not necessarily indicate the formation of equivalence classes \citep{SaundersGreen1992, Grisante_Tomanari2024, AlonsoAlvarez2023}.

The limited ability of ANN models to demonstrate equivalence class formation can be attributed to a fundamental discrepancy between their core design principles \citep{alpaydin2020, Goodfellow2016, vaswani2017attention, bahdanau2016attention} and the relational demands of stimulus equivalence tasks. Typically, supervised learning in ANN architectures prioritizes the establishment of discriminative response patterns by extracting and processing stimulus features within a defined feature space.

While these models are proficient at extending learned discriminations to novel stimuli sharing similar feature profiles, the generalization of functional relation is not a primary objective or a target value within their optimization processes. Consequently, the algorithmic optimization for feature discrimination may result in a functional mismatch with the demands of equivalence tasks, which necessitate responding based on derived relations rather than shared physical properties. This suggests that while ANNs can effectively acquire the trained conditional discriminations of an MTS procedure, their inherent bias towards feature-based learning may not readily facilitate the emergent relational responding that defines stimulus equivalence. 

Analyzing feature space representations \citep{Yu2015, Islam2024} provides a valuable methodology for further examining the nature of stimulus organization within these models, particularly to determine  whether it is based on acquired relational responding or surface feature similarities. For instance, \citet{TovarTorresChavez2012} employed clustering techniques and demonstrated that stimuli from the same equivalence class clustered within a reduced-dimensional feature space.

The current methodology extends beyond prior models by addressing limitations in encoding procedures, expanding the range of experimental conditions, and manipulating negative comparison selection. Previous simulations employing FFNs \citep{BarnesHampson1993, Lyddy2007, tovar_westerman_2017, Ninness2018, VernucioDebert2016} utilized fixed-position encoding schemes, wherein each stimulus was represented by a bit in a specific input vector location \citep{Tovar2023}. This encoding strategy rendered baseline trials and symmetry tests computationally indistinguishable due to identical encoded representations, thereby precluding the valid assessment of reflexivity. To address this constraint, we implemented the one-hot encoding method, as applied by \citet{CarrilloBetancort2023}, which permits stimulus position variation and consequently enables the evaluation of reflexivity, symmetry, and transitivity.
 
By incorporating 18 distinct experimental conditions, this study broadens the investigation of SE in computational models, exceeding the scope of many prior studies. This expanded design enables a more comprehensive analysis of the factors influencing agent performance. A key methodological innovation of this research is the manipulation of negative comparison selection, enabling a comparative analysis of standard and biased TS.

The present results are consistent with the findings of \citet{CarrilloBetancort2023, CarrilloBetancort2024}, who also employed a biased selection of negative comparisons, yet reveal divergent performance patterns under standard MTS selection. The observed performance differences between standard and biased negative comparison selection in our study indicate that their findings may not be directly comparable to typical MTS experiments. Nevertheless, both the current study and the aforementioned investigations converge on the role of reject relations in SE simulations, their impact on agent performance, and the consistent observation that connectionist agents do not fully exhibit SE.

Considering our findings in these simulations, prior models may have exhibited response patterns based on strategies other than equivalence class formation, although this was not explicitly examined or controlled in those studies. In contrast to RELNET's simulations performed by \citet{BarnesHampson1993} and \citet{Lyddy2007}, our study revealed that agents demonstrated deficits in reflexivity and generalization. Equivalence-like behavior in RELNET simulations may have been influenced by its architectural features, potentially obscuring the underlying learning mechanisms. Furthermore, our findings diverge from those of \citet{TovarTorresChavez2012} and \citet{VernucioDebert2016} with respect to trial encoding and the training and testing procedures for SE. That our simulations, with distinct encoding procedures and manipulations of reject relations, yielded different results underscores the substantial impact of these methodological choices on simulation outcomes.

A notable discrepancy emerges when comparing the performance of the computational agents to that of humans with basic language repertoires in SE tasks. While variations in training parameters can modulate the efficiency of equivalence acquisition in humans, they rarely preclude it entirely. The agents' persistent difficulty with the OTM structure contrasts with findings in human studies, wherein OTM structures can be as effective or slightly more effective than MTO structures, particularly with adults \citep{Arntzen2012, Ayres_PereiraVanessa2021}.

In humans, successful performance on reflexivity tests is considered a fundamental property of equivalence relations and a behavioral prerequisite for demonstrating symmetry and transitivity, serving as a foundational component of equivalence \citep{Sidman1994}. However, in the present study, the agents frequently failed to exhibit consistent and reliable reflexivity, even when demonstrating some success on other equivalence tests. This dissociation suggests that the agents' apparent success on some equivalence tests may result from strategies that bypass the requirement of stimulus identity, potentially analogous to generalized conditional responding observed in humans \citep{Plazas2024}. This discrepancy highlights aspects of human behavior that remain inadequately captured by current connectionist architectures.

The discrepancy between offline learning in our simulations and online human learning likely contributes to observed performance differences in stimulus equivalence tasks. Human learning's gradual emergence \citep{SaundersGreen1992}, sensitivity to trial order \citep{GreenSaunders1998}, and alignment with reinforcement learning principles \citep{Sutton2018Reinforcement, Mofrad2020, Silver2021RL} contrast with the holistic pattern recognition and batch processing of our models. Incorporating online learning mechanisms into future simulations is crucial for more accurately replicating human learning dynamics and achieving human-like equivalence behavior.

Simulations typically utilize pre-instantiated stimuli, encoded prior to presentation, while humans dynamically process complex, multi-faceted sensory input. Humans actively segment and instantiate individual stimuli before engaging in matching tasks and demonstrating equivalence class formation, even beyond standard MTS procedures \citep{Grisante2013, VernucioDebert2016, Lantaya2018}. This simplification limits the complexity of input processing, impairing the model's ability to handle the variability and richness of real-world sensory data. Consequently, the model's generalizability to context-dependent experimental protocols is compromised. Moreover, this approach omits the active role of human perception in constructing and stabilizing stimuli, potentially excluding a critical component of equivalence class formation. Future research should prioritize integrating sophisticated sensory processing layers \citep{Goodfellow2016}, attention mechanisms \citep{bahdanau2016attention, vaswani2017attention}, temporal dynamics, and hierarchical RL to replicate dynamic stimulus instantiation.

The sensitivity of agent performance to manipulations of select and reject relations underscores the necessity of rigorous methodological control within SE research. The configuration of TS, relation type, and negative comparison selection significantly influenced agent behavior, leading us to conclude that, even in the absence of demonstrated SE, S- stimuli impact learning. This aligns with research consistently demonstrating these variables as critical determinants of learning and the formation of equivalence classes \citep{GreenSaunders1998, Arntzen2012, Plazas2016, Plazas2018, Grisante_Tomanari2024}.

Reject relations and reject control can serve as methodological confounds, impacting the interpretation of research findings. Our simulations demonstrate that the presence of reject relations and reject control in SE may introduce non-equivalence mechanisms that alter data measurements and, consequently, the interpretation of equivalence class formation. To mitigate these challenges, we directly manipulated the training environment, utilizing select-only and reject-only conditions with dummy stimuli. This methodological innovation allowed us to isolate the individual contributions of select and reject relations to equivalence class formation, a task often challenging within the standard MTS procedure.

The stability of agent performance within the OTM TS, observed across variations in relation type and negative comparison selection, indicates a diminished influence of reject control, providing a valuable methodological baseline for future human and computational research. Within the OTM structure, a single stimulus per class functions as the sample, concentrating both select relations with intra-class members and reject relations with extra-class members on this specific node. This sample node occupies a unique position, being the sole intra-class member directly exposed to inter-class information, a pattern consistent across standard and biased selection conditions. Conversely, the MTO and LS structures utilize $M-1$ samples, distributing reject information across a broader range of intra-class stimuli. This distribution is further amplified in the biased selection condition, where the quantity of reject information is increased.

For studies aiming to maximize training trials, particularly ANN simulations, biased negative comparison selection within the OTM TS may be advantageous. Conversely, standard negative comparison selection is more appropriate for studies employing LS or MTO TS. However, human learning may differ significantly, and reject control may not be merely a ``confounding'' factor but an integral component of equivalence class formation. The role of reject control in human equivalence formation and testing remains a subject of ongoing debate \citep{Sidman1994, CarriganSidman1992, Arntzen2012, Plazas2016, Plazas2018, Perez_Tomanari_2020}, and the complex interplay between training structures and the type of stimulus control established during training is not fully understood. Future research is essential to delineate these nuanced interactions.
 
To further investigate the role of reject relations and agent response patterns, we conducted supplementary simulations using a sequential training and testing procedure. Specifically, we trained all baseline pairs with a single sample stimulus and subsequently evaluated all reflexivity, symmetry, and transitivity test trials. This process was iterated for each sample stimulus. The results demonstrated that agents responded accurately to all test trials involving the trained sample stimulus. However, they failed to respond accurately to trials involving stimuli used exclusively as comparisons during baseline training. This outcome reinforces our hypothesis that agent response patterns are primarily controlled by reject relations. The agents' inability to generalize to stimuli not used as samples during training further suggests the absence of equivalence class formation. This result also elucidates the performance patterns observed within the OTM structure.

The direct manipulation of reject relations, as implemented in this study, represents a novel experimental approach that, to our knowledge, has not yet been replicated with human participants. This gap between computational and behavioral research presents a significant avenue for future investigation \citep{Tovar2023}. The experimental conditions detailed herein can serve as a template for designing analogous human studies, facilitating a more nuanced examination of reject relations within SE. Extending this research to human subjects would not only validate the computational findings but also deepen our understanding of the behavioral mechanisms underlying equivalence class formation.

By preventing the formation of either select or reject relations during training and assessing equivalence class formation based solely on the other relation, our approach contrasts with novel stimulus tests. Novel stimulus tests introduce novelty during the test phase to probe existing stimulus control \citep{CarriganSidman1992, GreenSaunders1998, Boldrin2022}. This methodological divergence minimizes potential novelty bias during testing, enabling a more direct assessment of how training conditions influence equivalence class formation. 

This research presents strengths that enhance its contribution to the field of SE. First, the study addresses the theoretically significant and debated role of reject relations in SE through computational modeling. Second, the varied experimental conditions facilitate a comprehensive analysis of factors influencing agent performance, strengthening the study's ability to draw robust conclusions regarding the impact of reject relations. Notably, the manipulation of negative comparison selection, replicating prior ``biased selection'' findings \citep{CarrilloBetancort2024}, reveals distinct performance patterns under standard MTS selection. Third, the use of a probabilistic agent as an associative learning benchmark provides a critical control for interpreting equivalence test results, enabling the assessment of whether agents rely on associative learning of select and reject relations. Finally, the employment of diverse, advanced computational models (BERT, GPT, FFN) allows for the comparative analysis of different learning algorithms' capacity to model SE complexities.

We acknowledge several limitations in our study. First, the use of a single simulation run per condition limits the generalizability of our findings and raises the possibility that results are attributable to chance or run-specific factors, as variations in random weight initialization could yield different outcomes. Second, hyperparameters for the transformer models (BERT and GPT) were not optimized for this task, a potential limitation given neural network performance's sensitivity to these settings. Third, while informative, comparisons with a probabilistic agent may not encompass all potential non-SE strategies employed by the neural networks. The networks may utilize alternative associative strategies beyond those modeled by the probabilistic agent. Finally, the study did not analyze the networks' internal representations. Examination of these representations could offer valuable insights into whether equivalence processes occur, even if not overtly expressed in the simulations' output.

Future research directions could enhance ANN models for simulating SE and elucidate the underlying learning mechanisms. While systematic hyperparameter tuning and upscaling current models may improve performance on equivalence tasks, neither guarantees the demonstration of equivalence class formation. Beyond architectural enhancements, exploring alternative learning algorithms and their sensitivity to reject relations is warranted. Reinforcement learning-based algorithms, as demonstrated by \citet{Mofrad2020, Mofrad2021}, present a promising approach. These algorithms, grounded in behavioral theory \citep{Sutton2018Reinforcement, Silver2021RL}, may more accurately model the dynamic nature of human equivalence learning. The current study's findings, which highlight limitations in existing architectures, provide a clear impetus for developing novel models or training protocols capable of simulating SE. Further investigations could also explore online learning paradigms and stimulus instantiation from complex inputs. Finally, the focus on reject relations offers a specific and valuable avenue for future research.

\section{Conclusion}
This study investigated the influence of reject relations on equivalence class formation in computational models (FFN, BERT, GPT) across varying training structures, relation types, and negative comparison selection. Our findings provide limited support for equivalence class formation across a wide range of simulation parameters. While proficient in baseline discriminations, the agents relied on associative learning based on select and reject relations, rather than demonstrating SE. Performance was highly sensitive to reject information availability, with agents failing to generalize when S- relations were absent and performing comparably to a probabilistic agent. The divergence between agent and human performance underscores limitations in current models' ability to replicate human-like SE. The consistent failure of reflexivity, inconsistent performance on equivalence tests across conditions, and the absence of successful OTM performance further suggest a lack of equivalence class formation.

The manipulation of reject relations significantly influenced agent performance, indicating that an overemphasis on S- during training, particularly in biased negative comparison conditions, enabled accurate responding on equivalence tests through mechanisms other than derived relational responding, such as associative learning or exclusion. Training structure also played a critical role; LS and MTO structures yielded more instances of higher test performance compared to the OTM structure, regardless of relation type or negative comparison strategy. Furthermore, the performance of the connectionist models often mirrored that of the probabilistic agent, especially in conditions with a higher density of reject relations, suggesting that these sophisticated architectures may have relied on similar associative strategies rather than exhibiting the emergent properties characteristic of SE. The absence of explicit reject relations during training, exemplified by the select-only condition, generally resulted in poor test performance, indicating the importance of these relations for demonstrating equivalence. This study’s manipulation of negative comparison selection provides further insight into how the balance of S+ and S- relations during training can impact subsequent equivalence test performance.

Our simulations suggest that achieving accurate performance on equivalence tests in computational agents does not necessarily indicate the formation of equivalence classes. An over-reliance on reject relations during training may facilitate responding based on exclusion or associative learning, mimicking some aspects of equivalence without the underlying derived relational responding. These results emphasize the importance of considering reject relations in computational models of equivalence and suggest that current ANN architectures may not fully capture human SE. The innovations implemented here advance computational modeling of equivalence, offering a more nuanced understanding of factors influencing class formation in agents. Future research exploring newer models, hyperparameter tuning, specialized algorithms, and reinforcement learning-based algorithms promises to further advance this field.

{
\small
\bibliographystyle{apalike}
\bibliography{references} 
}

\newpage
\appendix
\renewcommand{\thefigure}{S\arabic{figure}}
\setcounter{figure}{0}

\section*{Appendix}
\subsection*{Detailed Simulation Results}\label{Apx_results}
This appendix presents detailed results from all 72 simulations. Table \ref{tab:performance_complete} provides a comprehensive overview of the performance metrics for all 72 simulations conducted in this study. Specifically, it presents the correct selection rates for baseline training, reflexivity tests, symmetry tests, and transitivity tests. Results are categorized by training structure (LS, MTO, OTM), relation type (Select-Reject, Select-Only, Reject-Only), negative comparison selection type (Standard, Biased), and the specific agent used in the simulation (BERT, FFN, GPT, Probabilistic). Values in boldface indicate performance that met or exceeded the mastery criterion of 0.9, allowing for a clear identification of simulations where agents demonstrated successful performance on the respective tests.

\begin{longtable}{@{}lllllrrrr@{}}
\caption{Performance metrics (correct selection rate) for baseline (Base) training, reflexivity (Refl) tests, symmetry (Symm) tests, and transitivity (Tran) tests across all 72 simulations (Simul), categorized by training structure (TS), relation type (Rel), negative comparison (NegC) selection type, and agent. Values in bold indicate performance at or above the mastery criterion of 0.9.}
\label{tab:performance_complete}\\
\toprule
\multirow{2}{*}{\textbf{Simuli}} & \multirow{2}{*}{\textbf{TS}} & \multirow{2}{*}{\textbf{Rel}} & \multirow{2}{*}{\textbf{NegC}} & \multirow{2}{*}{\textbf{Agent}} & \multicolumn{4}{l}{\textbf{Correct selection rate}} \\
\cmidrule(lr){6-9}
& & & & & \textbf{Base} & \textbf{Refl} & \textbf{Symm} & \textbf{Tran} \\
\midrule
\endhead
\bottomrule
\multicolumn{9}{r}{\footnotesize Continued on next page} \\
\endfoot
\bottomrule
\endlastfoot
1 & LS & Sel-Rej & Std & BERT & \textbf{1,00} & 0,09 & 0,56 & 0,30 \\
2 & LS & Sel-Rej & Std & FFN & \textbf{1,00} & 0,30 & 0,33 & 0,34 \\
3 & LS & Sel-Rej & Std & GPT & \textbf{0,99} & 0,04 & 0,62 & 0,33 \\
4 & LS & Sel-Rej & Std & Prob & \textbf{1,00} & 0,25 & 0,32 & 0,33 \\
5 & LS & Sel-Rej & B(S-) & BERT & \textbf{1,00} & \textbf{0,99} & \textbf{0,97} & \textbf{0,95} \\
6 & LS & Sel-Rej & B(S-) & FFN & \textbf{1,00} & 0,89 & 0,87 & 0,87 \\
7 & LS & Sel-Rej & B(S-) & GPT & \textbf{1,00} & 0,72 & \textbf{0,96} & 0,84 \\
8 & LS & Sel-Rej & B(S-) & Prob & \textbf{1,00} & 0,88 & \textbf{0,95} & 0,86 \\
9 & LS & Sel & Std & BERT & \textbf{1,00} & 0,32 & 0,36 & 0,33 \\
10 & LS & Sel & Std & FFN & \textbf{1,00} & 0,36 & 0,35 & 0,35 \\
11 & LS & Sel & Std & GPT & \textbf{0,99} & 0,08 & 0,15 & 0,14 \\
12 & LS & Sel & Std & Prob & \textbf{1,00} & 0,26 & 0,30 & 0,35 \\
13 & LS & Sel & B(S-) & BERT & \textbf{1,00} & 0,28 & 0,32 & 0,32 \\
14 & LS & Sel & B(S-) & FFN & \textbf{1,00} & 0,35 & 0,35 & 0,33 \\
15 & LS & Sel & B(S-) & GPT & \textbf{1,00} & 0,03 & 0,06 & 0,03 \\
16 & LS & Sel & B(S-) & Prob & \textbf{1,00} & 0,28 & 0,23 & 0,36 \\
17 & LS & Rej & Std & BERT & \textbf{1,00} & 0,33 & 0,30 & 0,34 \\
18 & LS & Rej & Std & FFN & \textbf{1,00} & 0,30 & 0,33 & 0,34 \\
19 & LS & Rej & Std & GPT & \textbf{1,00} & 0,27 & 0,33 & 0,33 \\
20 & LS & Rej & Std & Prob & \textbf{1,00} & 0,31 & 0,38 & 0,30 \\
21 & LS & Rej & B(S-) & BERT & \textbf{1,00} & 0,33 & 0,33 & 0,32 \\
22 & LS & Rej & B(S-) & FFN & \textbf{1,00} & 0,36 & 0,36 & 0,35 \\
23 & LS & Rej & B(S-) & GPT & \textbf{1,00} & 0,34 & 0,34 & 0,34 \\
24 & LS & Rej & B(S-) & Prob & \textbf{1,00} & 0,85 & 0,88 & 0,85 \\
25 & MTO & Sel-Rej & Std & BERT & \textbf{1,00} & 0,20 & 0,04 & 0,22 \\
26 & MTO & Sel-Rej & Std & FFN & \textbf{1,00} & 0,34 & 0,33 & 0,35 \\
27 & MTO & Sel-Rej & Std & GPT & \textbf{1,00} & 0,04 & 0,82 & 0,04 \\
28 & MTO & Sel-Rej & Std & Prob & \textbf{1,00} & 0,21 & 0,47 & 0,22 \\
29 & MTO & Sel-Rej & B(S-) & BERT & \textbf{1,00} & 0,34 & 0,30 & 0,36 \\
30 & MTO & Sel-Rej & B(S-) & FFN & \textbf{1,00} & 0,84 & 0,34 & \textbf{0,93} \\
31 & MTO & Sel-Rej & B(S-) & GPT & \textbf{1,00} & 0,33 & 0,39 & 0,34 \\
32 & MTO & Sel-Rej & B(S-) & Prob & \textbf{1,00} & \textbf{0,93} & 0,23 & \textbf{1,00} \\
33 & MTO & Sel & Std & BERT & \textbf{1,00} & 0,31 & 0,30 & 0,31 \\
34 & MTO & Sel & Std & FFN & \textbf{1,00} & 0,33 & 0,34 & 0,34 \\
35 & MTO & Sel & Std & GPT & \textbf{1,00} & 0,32 & 0,31 & 0,35 \\
36 & MTO & Sel & Std & Prob & \textbf{1,00} & 0,18 & 0,33 & 0,34 \\
37 & MTO & Sel & B(S-) & BERT & \textbf{1,00} & 0,32 & 0,32 & 0,34 \\
38 & MTO & Sel & B(S-) & FFN & \textbf{1,00} & 0,33 & 0,35 & 0,36 \\
39 & MTO & Sel & B(S-) & GPT & \textbf{1,00} & 0,31 & 0,31 & 0,34 \\
40 & MTO & Sel & B(S-) & Prob & \textbf{1,00} & 0,44 & 0,37 & 0,30 \\
41 & MTO & Rej & Std & BERT & \textbf{0,99} & 0,31 & 0,27 & 0,32 \\
42 & MTO & Rej & Std & FFN & \textbf{1,00} & 0,33 & 0,33 & 0,32 \\
43 & MTO & Rej & Std & GPT & \textbf{1,00} & 0,36 & 0,32 & 0,33 \\
44 & MTO & Rej & Std & Prob & \textbf{1,00} & 0,43 & 0,23 & 0,36 \\
45 & MTO & Rej & B(S-) & BERT & \textbf{1,00} & 0,32 & 0,36 & 0,33 \\
46 & MTO & Rej & B(S-) & FFN & \textbf{1,00} & 0,34 & 0,30 & 0,36 \\
47 & MTO & Rej & B(S-) & GPT & \textbf{1,00} & 0,35 & 0,31 & 0,34 \\
48 & MTO & Rej & B(S-) & Prob & \textbf{1,00} & 0,89 & 0,45 & \textbf{1,00} \\
49 & OTM & Sel-Rej & Std & BERT & \textbf{1,00} & 0,06 & 0,03 & 0,07 \\
50 & OTM & Sel-Rej & Std & FFN & \textbf{1,00} & 0,30 & 0,34 & 0,31 \\
51 & OTM & Sel-Rej & Std & GPT & \textbf{1,00} & 0,02 & 0,64 & 0,03 \\
52 & OTM & Sel-Rej & Std & Prob & \textbf{1,00} & 0,42 & 0,48 & 0,30 \\
53 & OTM & Sel-Rej & B(S-) & BERT & \textbf{1,00} & 0,34 & 0,13 & 0,27 \\
54 & OTM & Sel-Rej & B(S-) & FFN & \textbf{1,00} & 0,44 & 0,31 & 0,32 \\
55 & OTM & Sel-Rej & B(S-) & GPT & \textbf{1,00} & 0,36 & 0,52 & 0,40 \\
56 & OTM & Sel-Rej & B(S-) & Prob & \textbf{1,00} & 0,35 & 0,28 & 0,32 \\
57 & OTM & Sel & Std & BERT & \textbf{1,00} & 0,33 & 0,30 & 0,35 \\
58 & OTM & Sel & Std & FFN & \textbf{1,00} & 0,30 & 0,33 & 0,33 \\
59 & OTM & Sel & Std & GPT & \textbf{1,00} & 0,34 & 0,35 & 0,33 \\
60 & OTM & Sel & Std & Prob & \textbf{1,00} & 0,22 & 0,20 & 0,31 \\
61 & OTM & Sel & B(S-) & BERT & \textbf{1,00} & 0,28 & 0,31 & 0,33 \\
62 & OTM & Sel & B(S-) & FFN & \textbf{1,00} & 0,32 & 0,34 & 0,32 \\
63 & OTM & Sel & B(S-) & GPT & \textbf{1,00} & 0,28 & 0,31 & 0,30 \\
64 & OTM & Sel & B(S-) & Prob & \textbf{1,00} & 0,24 & 0,33 & 0,36 \\
65 & OTM & Rej & Std & BERT & \textbf{1,00} & 0,11 & 0,26 & 0,17 \\
66 & OTM & Rej & Std & FFN & \textbf{1,00} & 0,34 & 0,33 & 0,33 \\
67 & OTM & Rej & Std & GPT & \textbf{1,00} & 0,41 & 0,40 & 0,37 \\
68 & OTM & Rej & Std & Prob & \textbf{1,00} & 0,38 & 0,38 & 0,31 \\
69 & OTM & Rej & B(S-) & BERT & \textbf{1,00} & 0,32 & 0,32 & 0,31 \\
70 & OTM & Rej & B(S-) & FFN & \textbf{1,00} & 0,36 & 0,31 & 0,30 \\
71 & OTM & Rej & B(S-) & GPT & \textbf{1,00} & 0,24 & 0,23 & 0,25 \\
72 & OTM & Rej & B(S-) & Prob & \textbf{1,00} & 0,43 & 0,28 & 0,35
\end{longtable}
{\footnotesize
Sel-Rej = Select-reject, Sel = Select-only, Rej = Reject-only, Standard = Std, Biased (S-) = B(S-), Prob = Probabilistic
}

\subsection*{Comparing trained select and reject relations across experimental conditions}\label{Apx_relations}
The figures in this appendix depict the trained relations for each experimental condition, according to table \ref{tab:experimental_conditions}. Rows represent the sample stimuli, and columns the comparison stimuli. Each colored cell indicates the pairs used during the training phase for the specific condition. Pairs used as baseline relations are colored blue, and pairs used as comparisons are colored red, representing the relation type information available to the agents: red signifies a reject relation (S-), and blue indicates a select relation (S+).

On both axes, stimuli were ordered first by class and then by member. Class 1 and its members (A through F) are presented first, followed by Class 2, and so on (A1, B1, C1, D1, ..., F4). Training structures exhibit different patterns of blue cells for class member stimuli. Select-only and reject-only conditions used dummy stimuli (Z\_11, Z\_12, Z\_13...), presented on the comparison axis (columns) after the class member stimuli. Patterns of reject relations with other class members are also presented as red cells.

\begin{figure*}[h]
    \centering
    \caption{LS (standard)}
    \includegraphics[height=0.4\textheight]{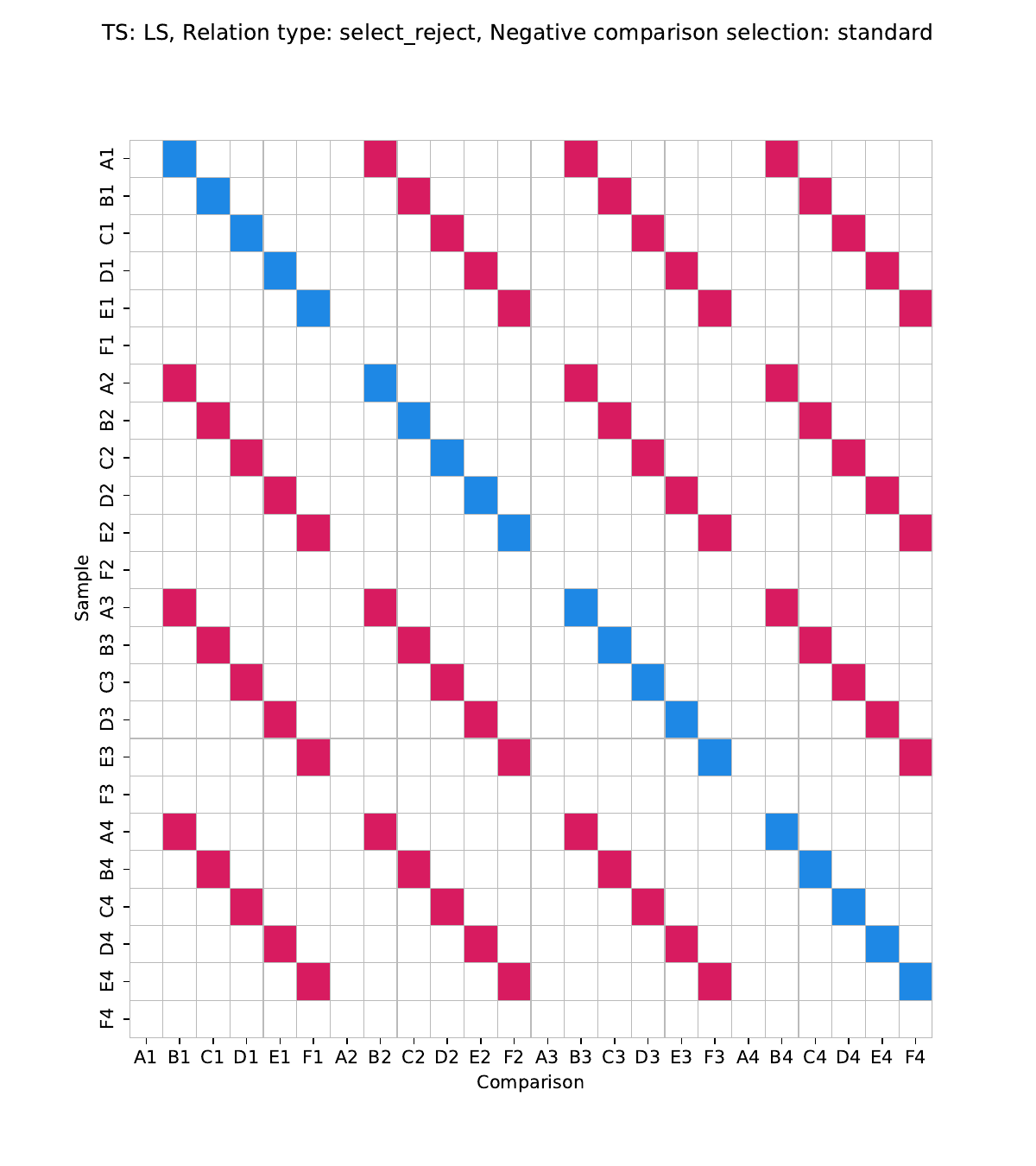}
    \label{fig:LS_standard}
\end{figure*}

\begin{figure*}[h]
    \centering
    \caption{LS, select-only with dummy stimuli}
    \includegraphics[height=0.4\textheight]{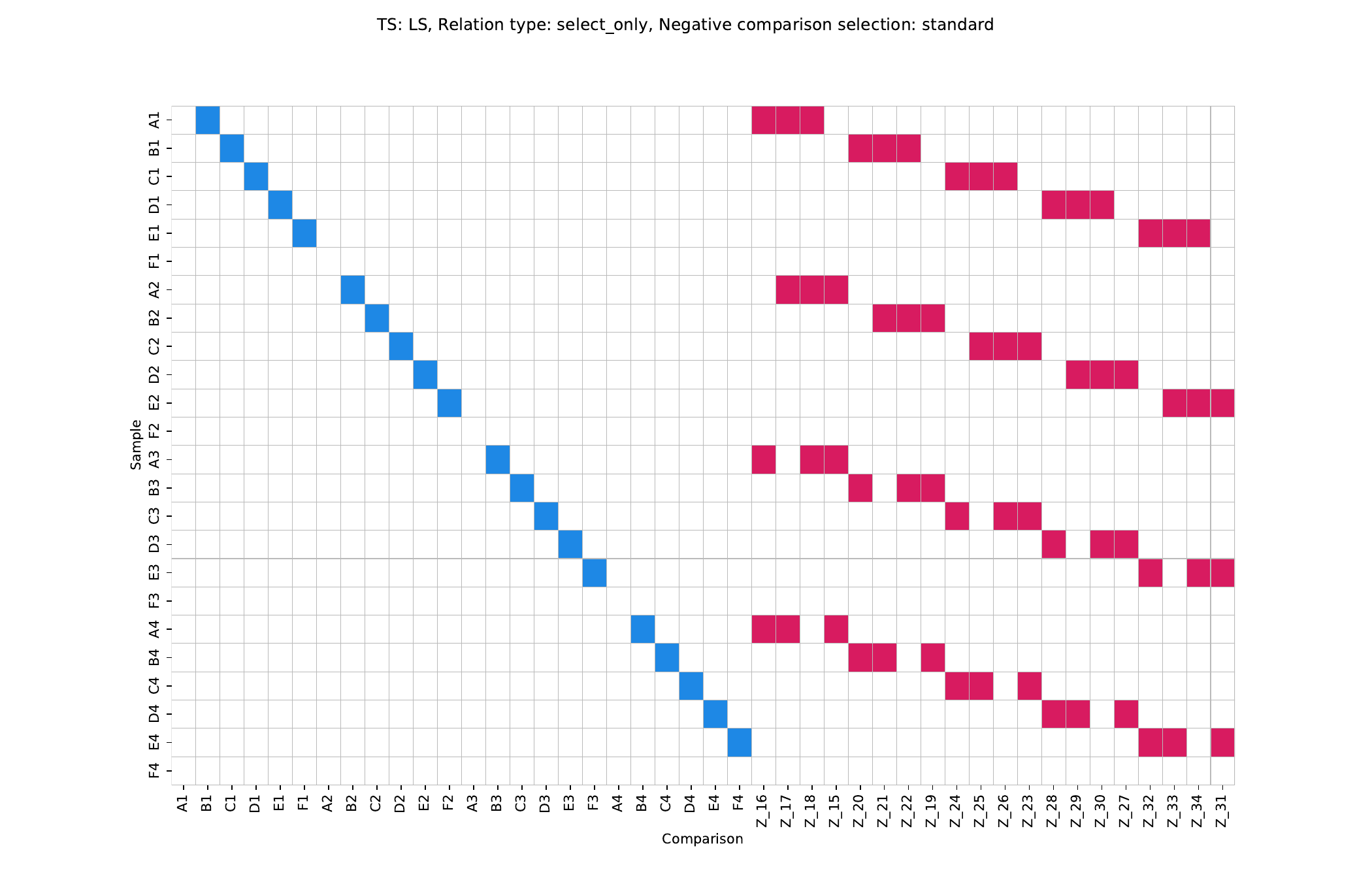}
    \label{fig:LS_select}
\end{figure*}

\begin{figure*}[h]
    \centering
    \caption{LS, reject-only with dummy stimuli}
    \includegraphics[height=0.4\textheight]{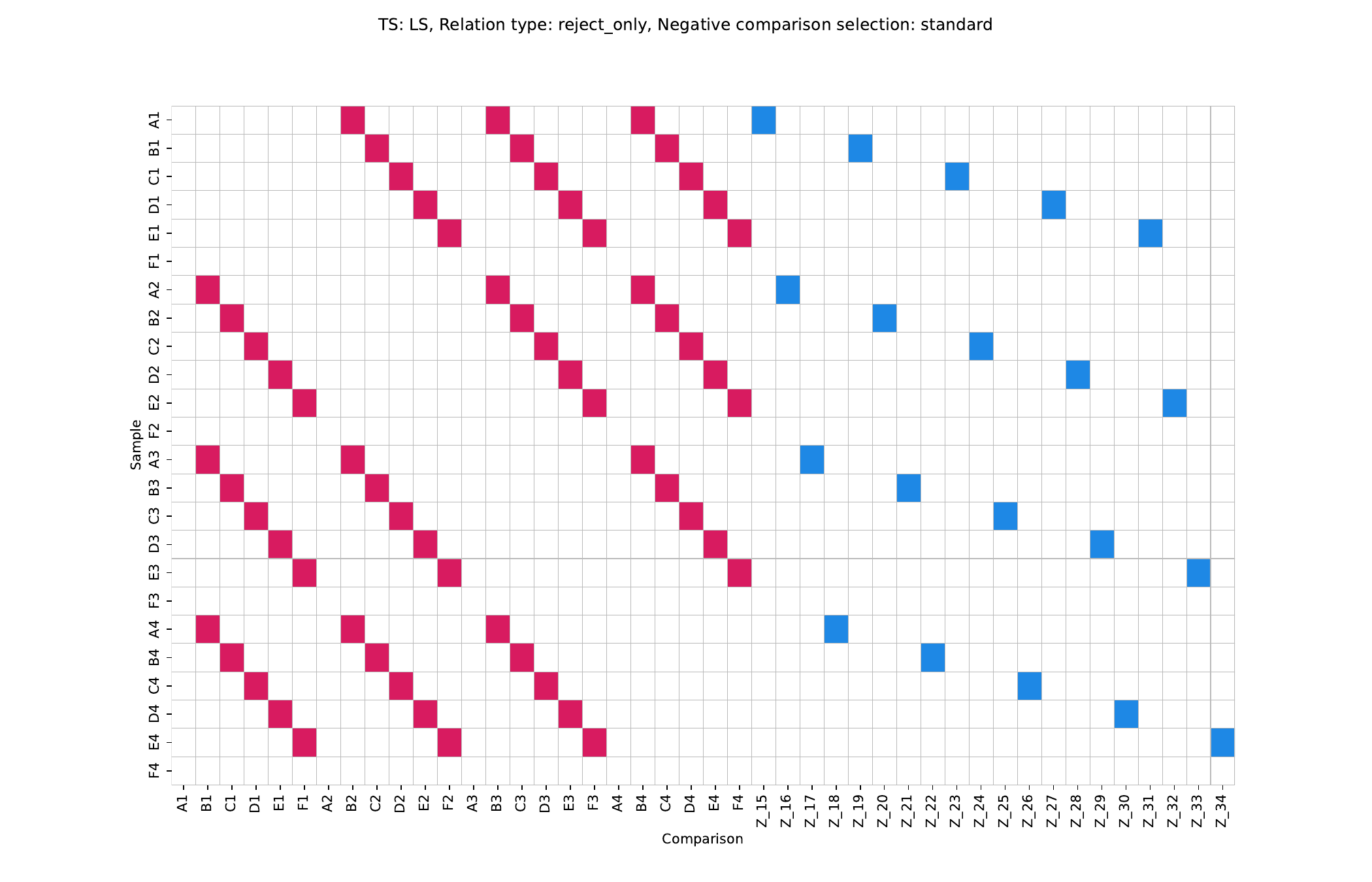}
    \label{fig:LS_reject}
\end{figure*}

\begin{figure*}[h]
    \centering
    \caption{LS biased}
    \includegraphics[height=0.4\textheight]{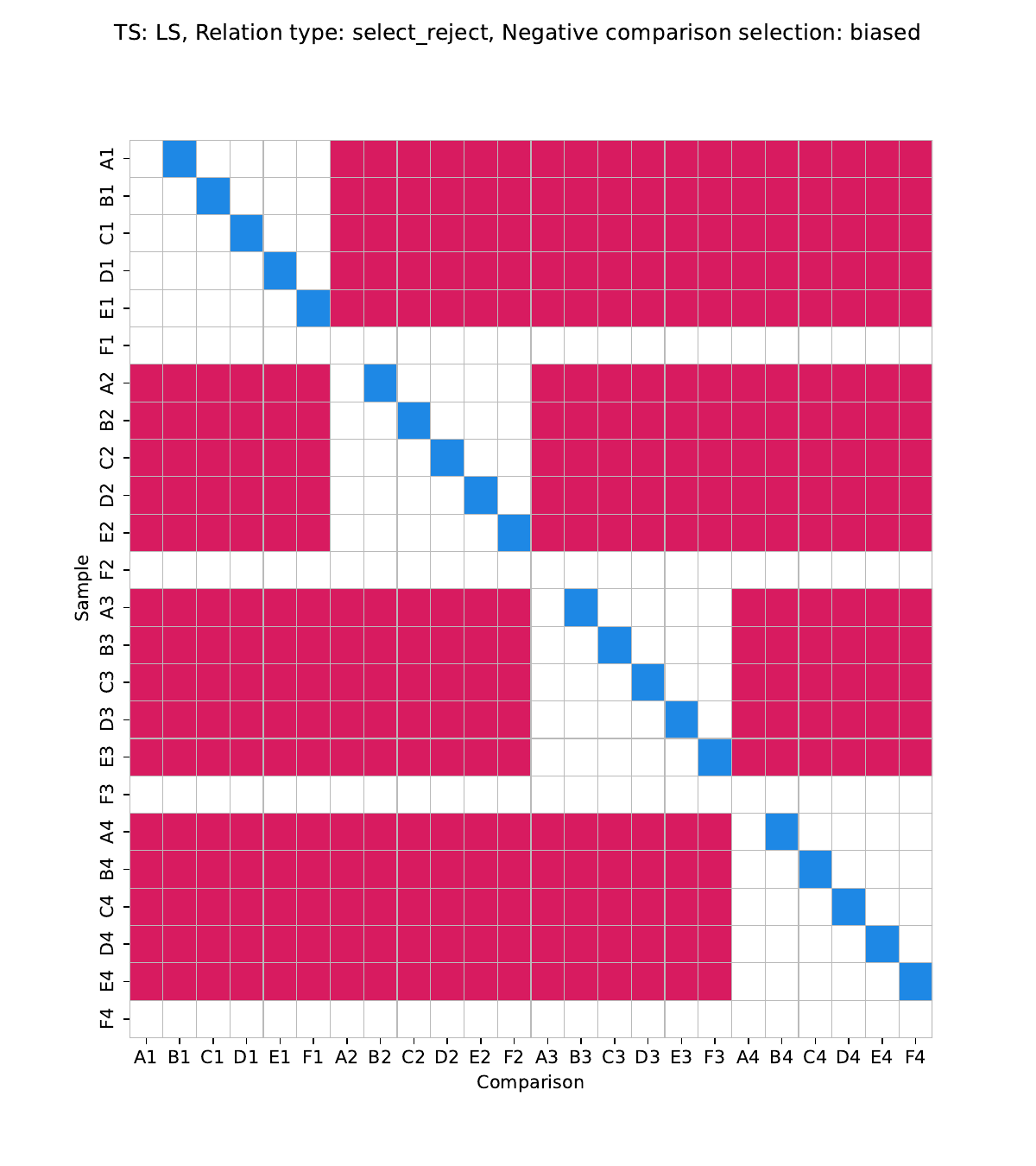}
    \label{fig:LS_biased}
\end{figure*}%

\begin{figure*}[h]
    \centering
    \caption{LS, select-only, biased, with dummy stimuli}
    \includegraphics[height=0.4\textheight]{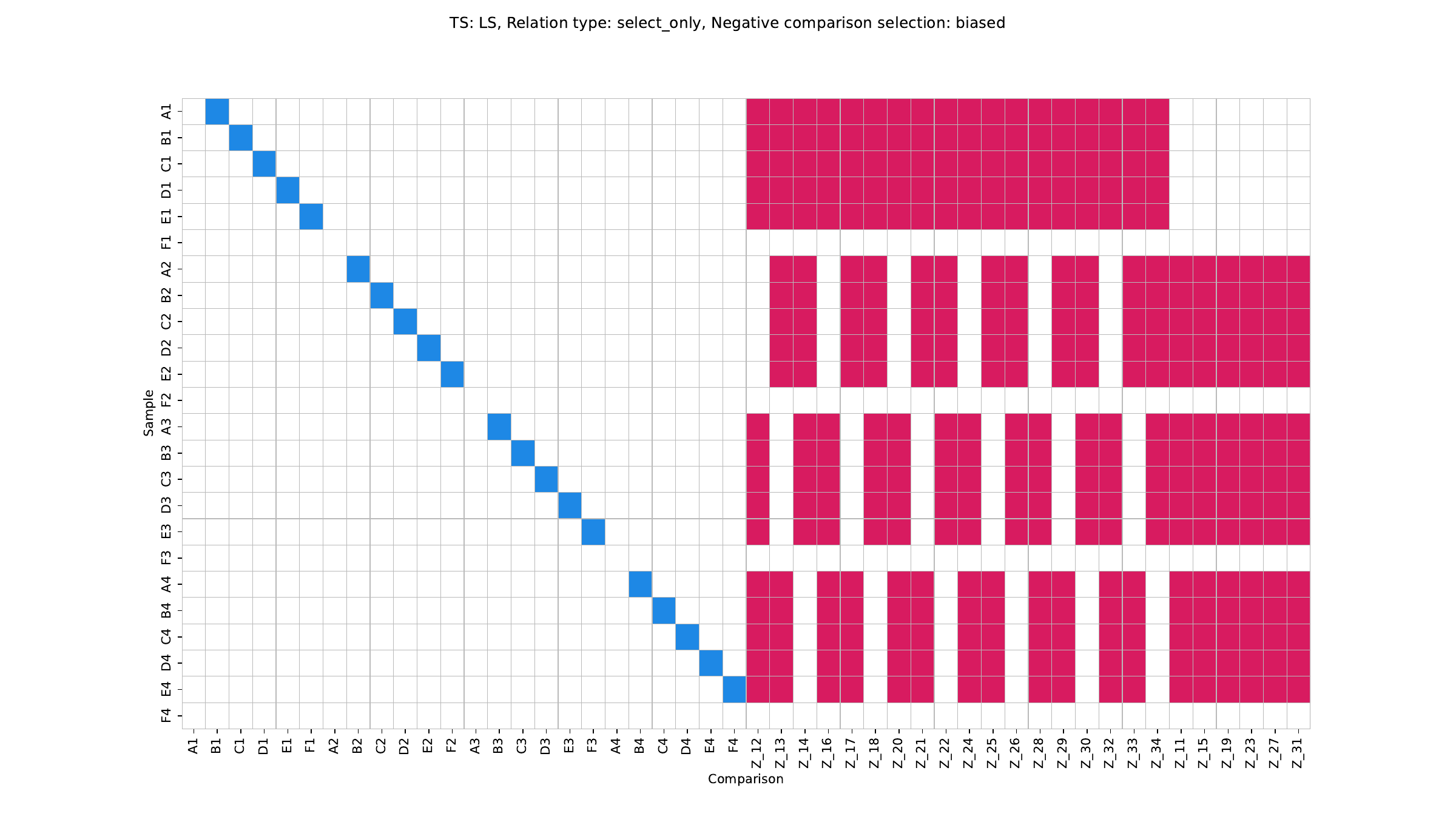}
    \label{fig:LS_select_biased}
\end{figure*}

\begin{figure*}[h]
    \centering
    \caption{LS, reject-only, biased, with dummy stimuli}
    \includegraphics[height=0.4\textheight]{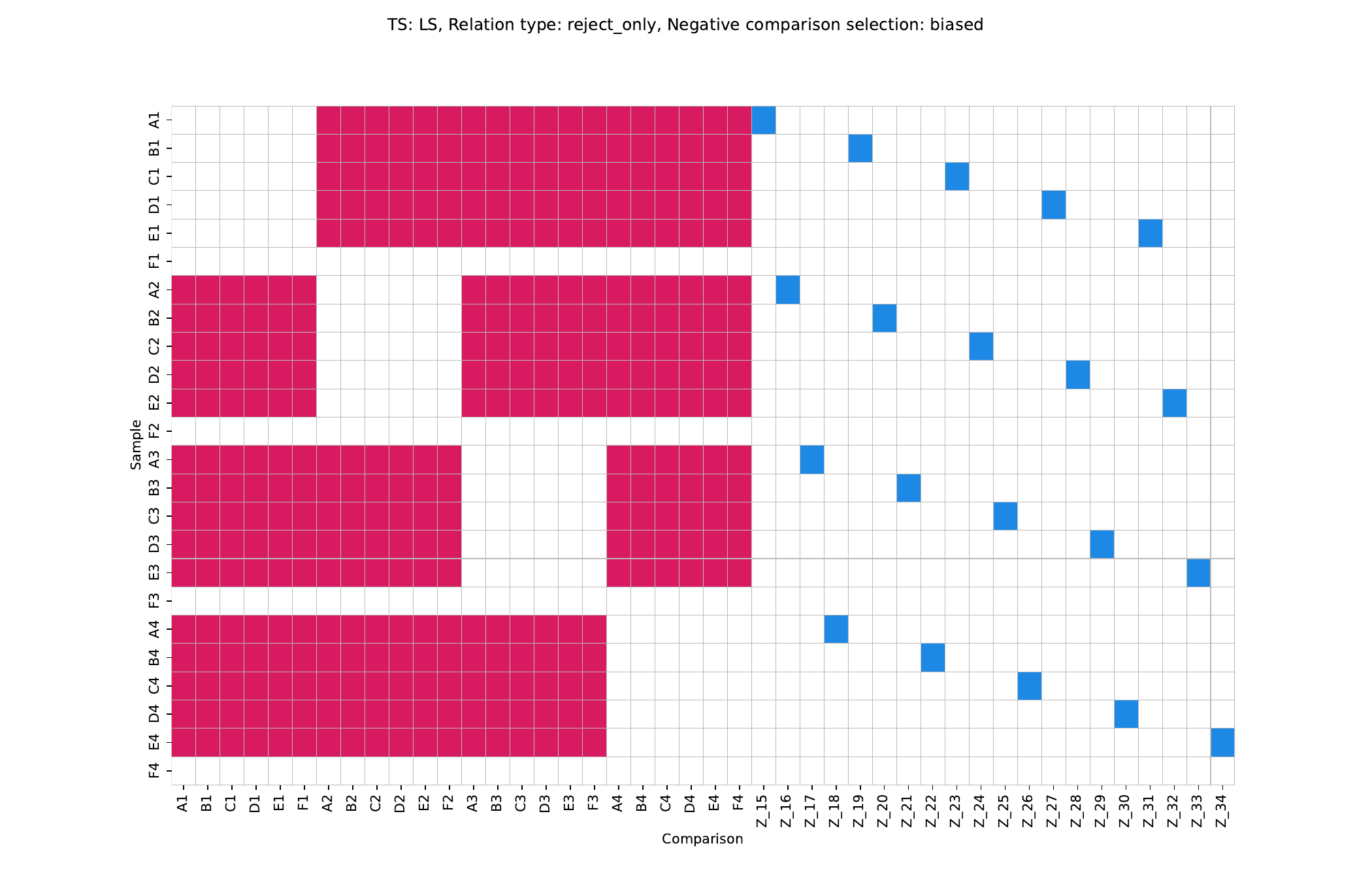}
    \label{fig:LS_reject_biased}
\end{figure*}

\begin{figure*}[h]
    \centering
    \caption{OTM (standard)}
    \includegraphics[height=0.4\textheight]{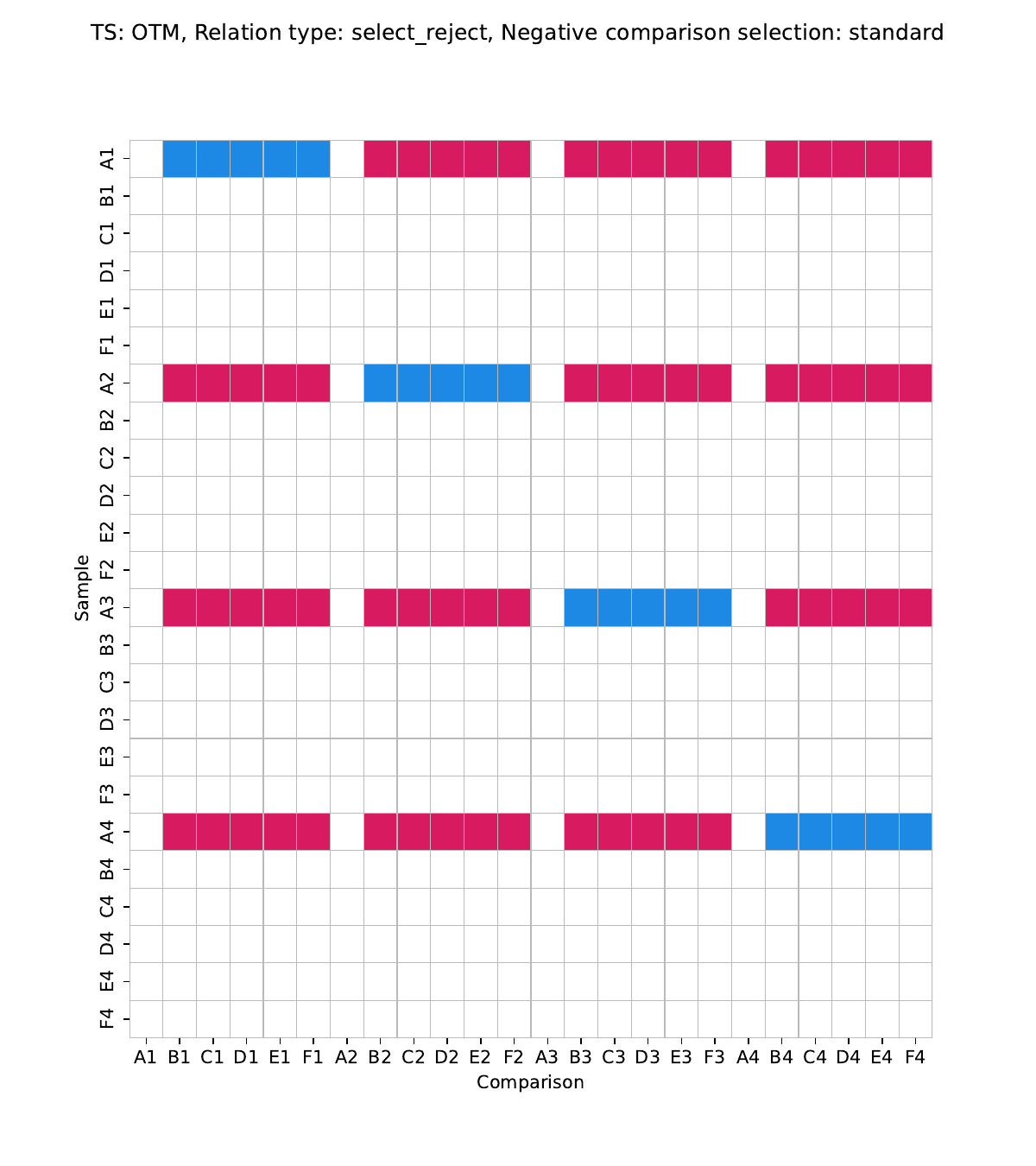}
    \label{fig:OTM_standard}
\end{figure*}

\begin{figure*}[h]
    \centering
    \caption{OTM, select-only, with dummy stimuli}
    \includegraphics[height=0.4\textheight]{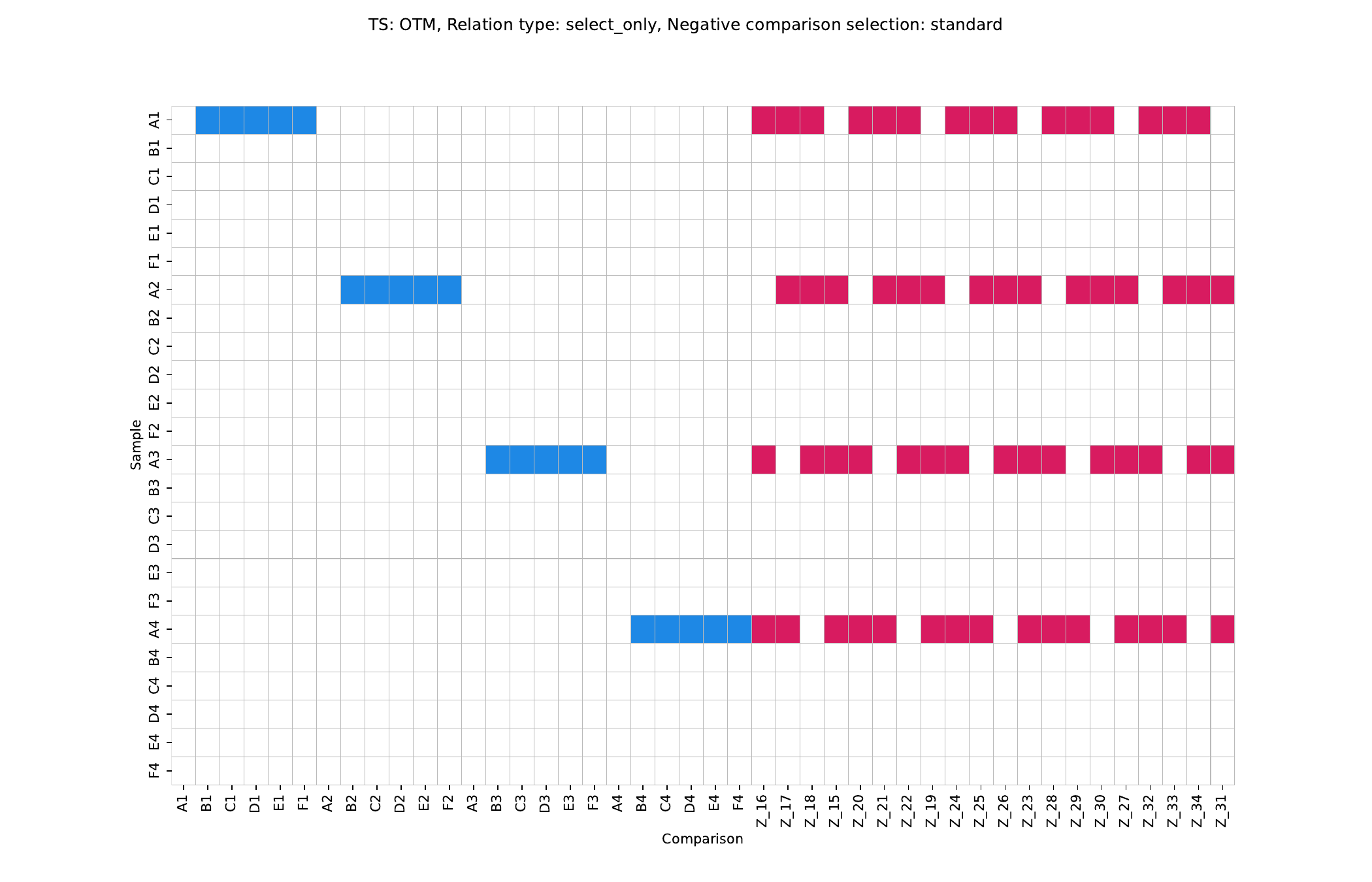}
    \label{fig:OTM_select}
\end{figure*}

\begin{figure*}[h]
    \centering
    \caption{OTM, reject-only, with dummy stimuli}
    \includegraphics[height=0.4\textheight]{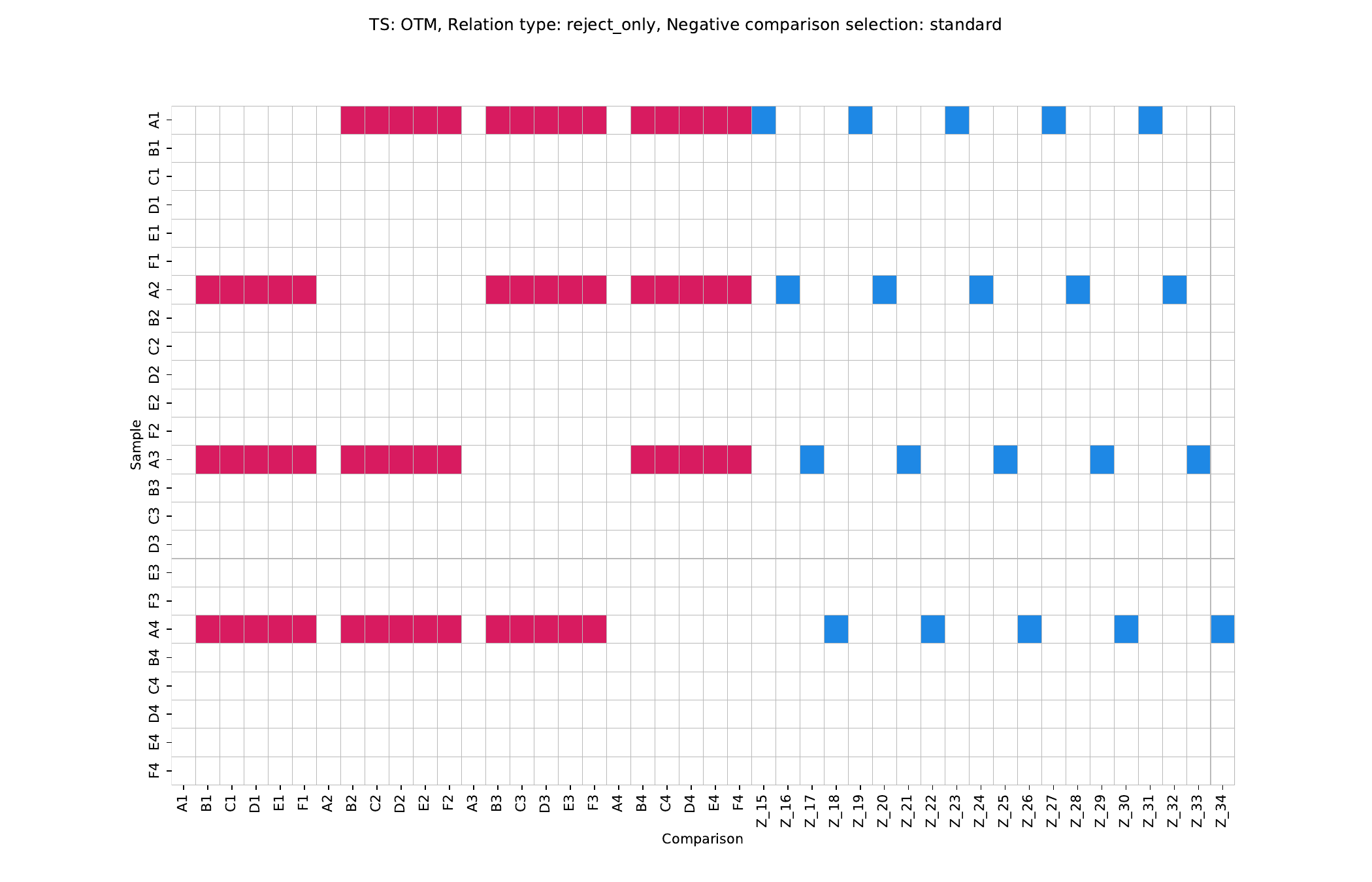}
    \label{fig:OTM_reject}
\end{figure*}

\begin{figure*}[h]
    \centering
    \caption{OTM biased}
    \includegraphics[height=0.4\textheight]{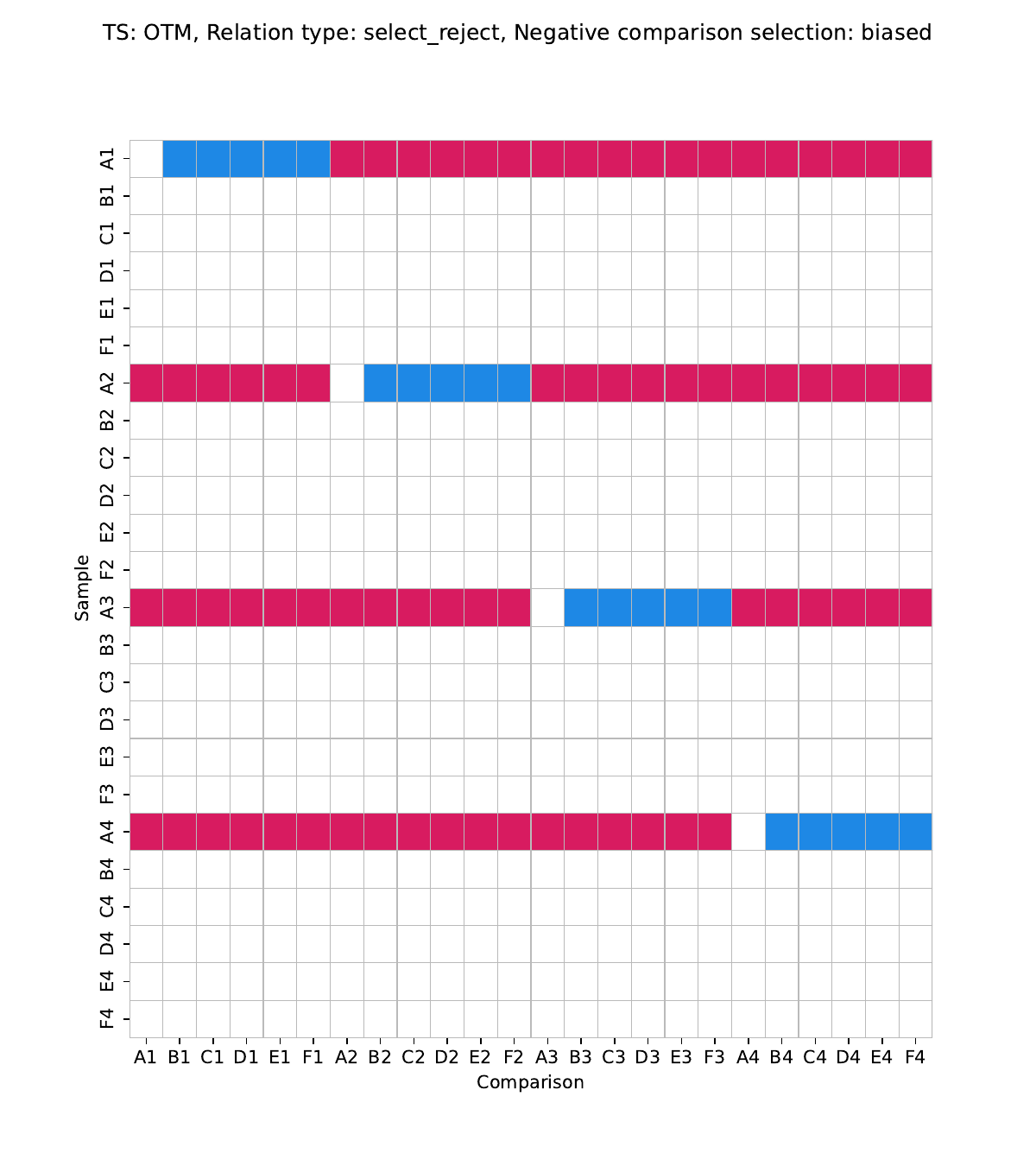}
    \label{fig:OTM_biased}
\end{figure*}

\begin{figure*}[h]
    \centering
    \caption{OTM, select-only, biased, with dummy stimuli}
    \includegraphics[height=0.4\textheight]{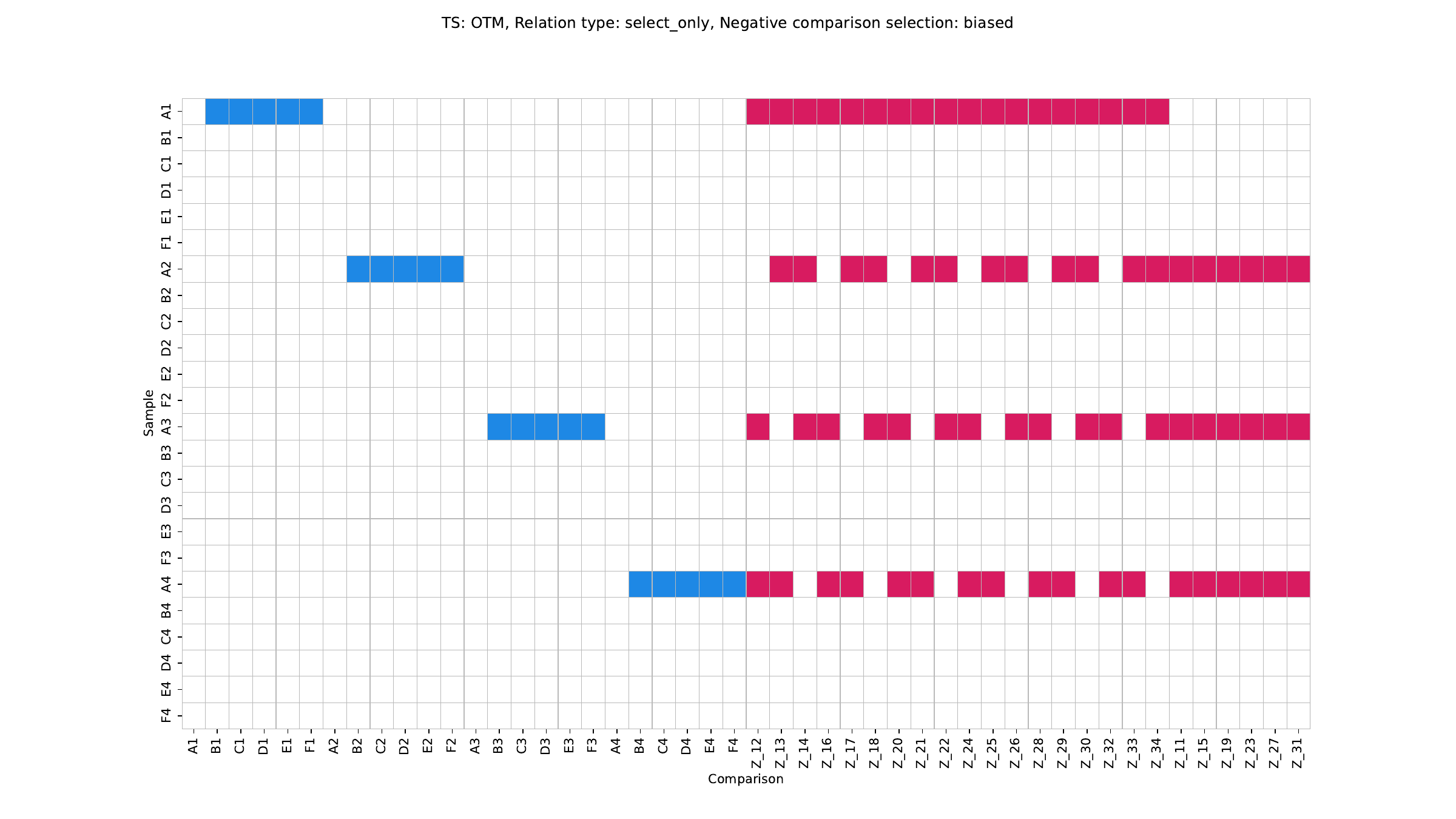}
    \label{fig:OTM_select_biased}
\end{figure*}

\begin{figure*}[h]
    \centering
    \caption{OTM , reject-only, biased, with dummy stimuli}
    \includegraphics[height=0.4\textheight]{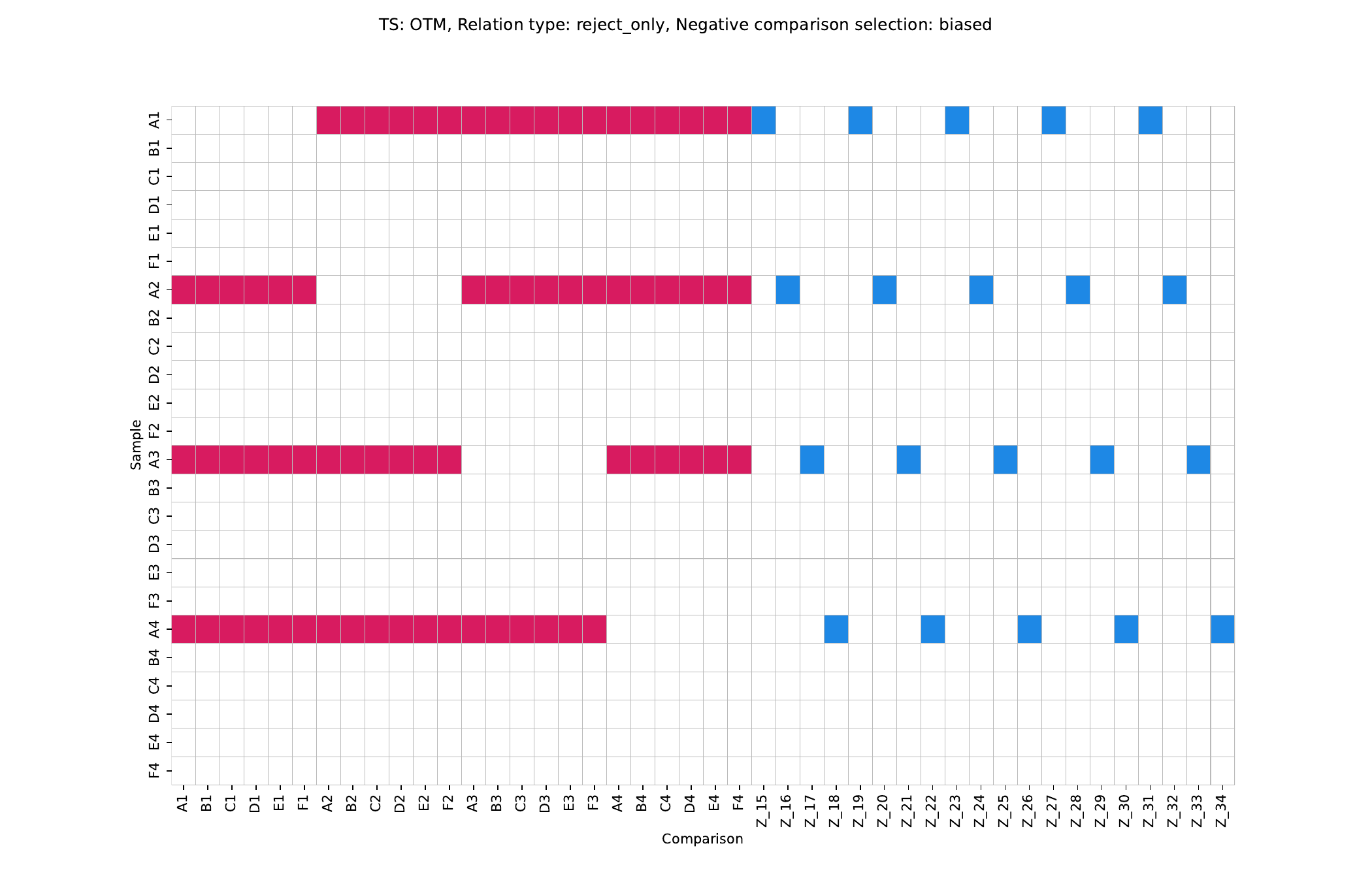}
    \label{fig:OTM_reject_biased}
\end{figure*}

\begin{figure*}[h]
    \centering
    \caption{MTO (standard)}
    \includegraphics[height=0.4\textheight]{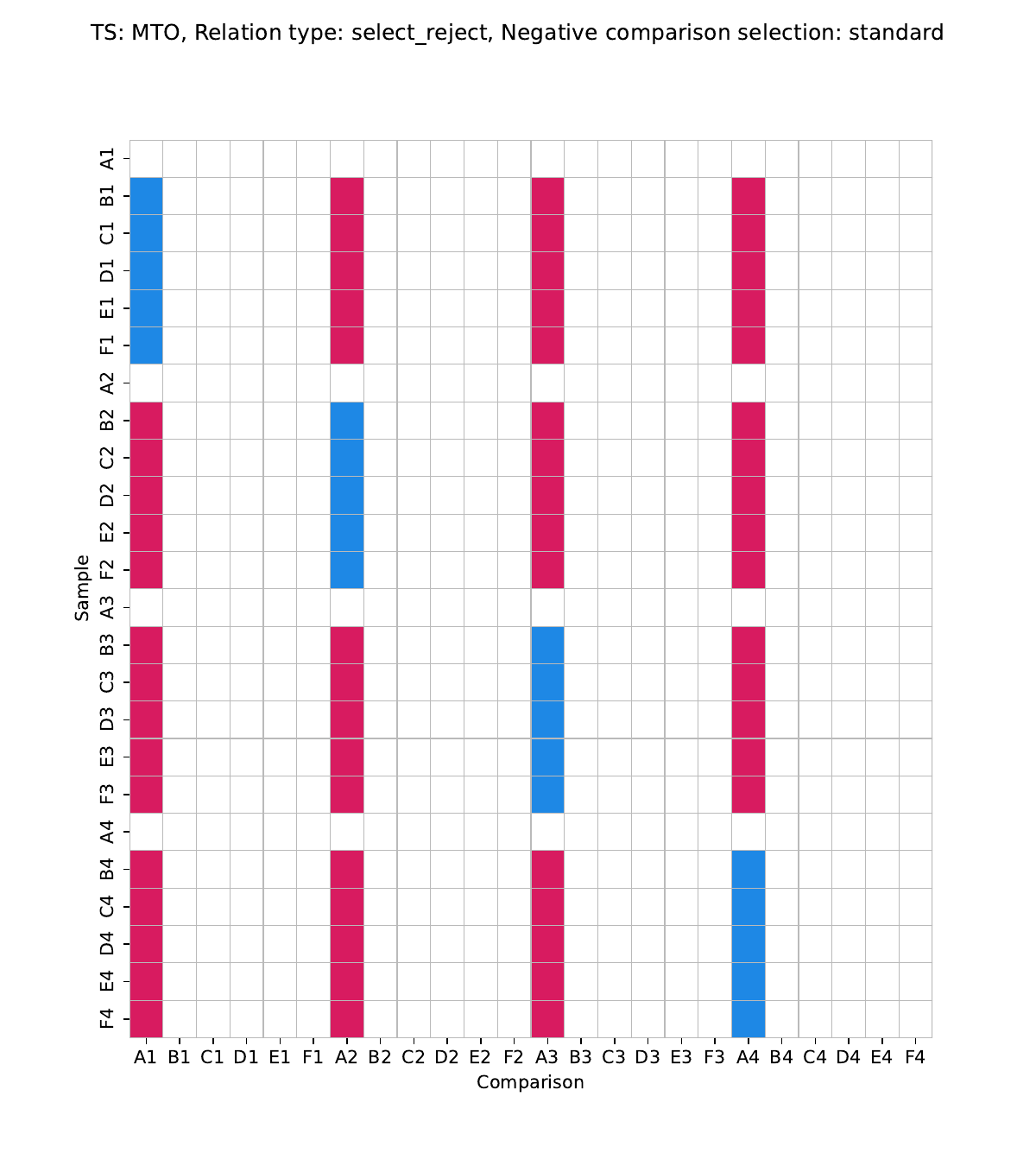}
    \label{fig:MTO_standard}
\end{figure*}

\begin{figure*}[h]
    \centering
    \caption{MTO, select-only, with dummy stimuli}
    \includegraphics[height=0.4\textheight]{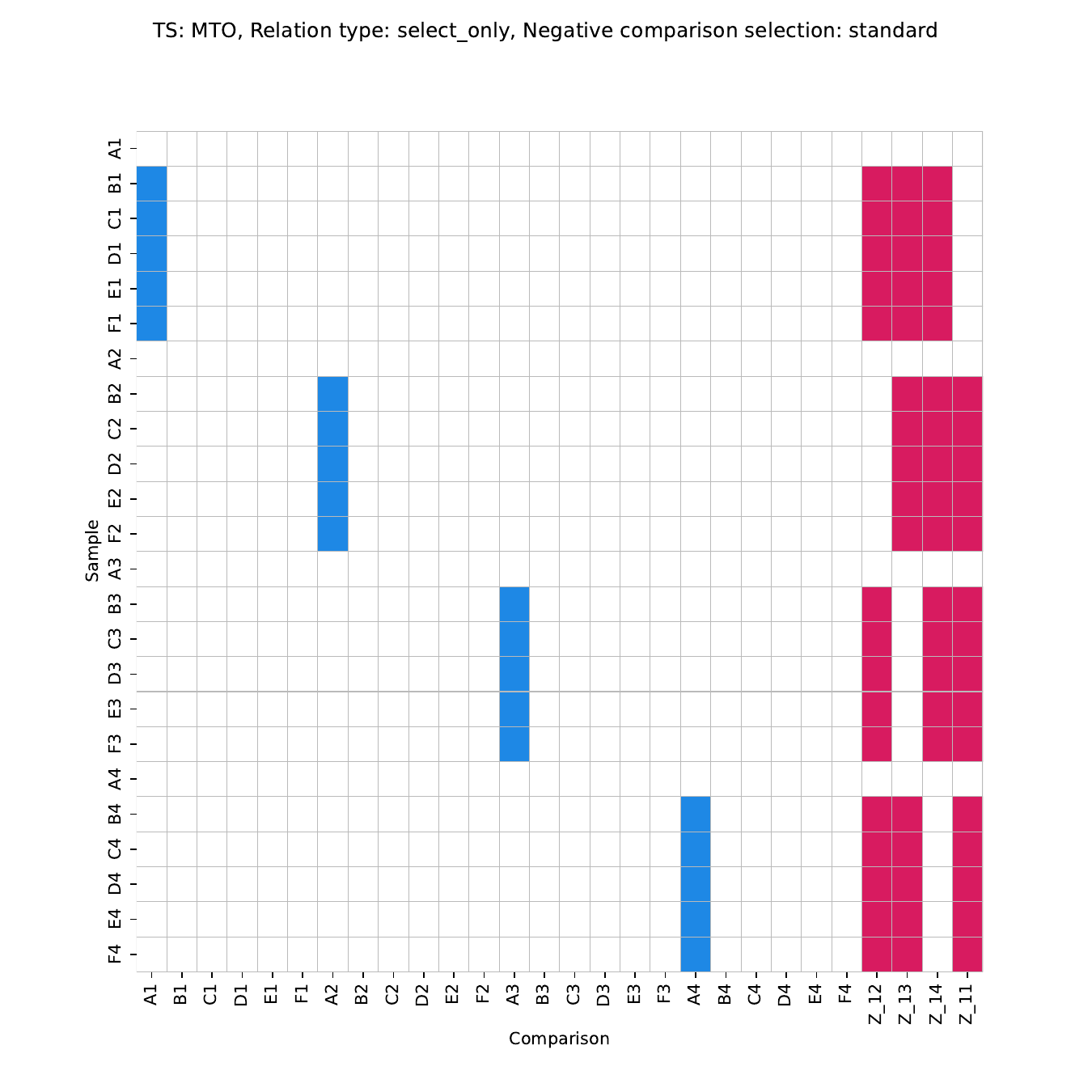}
    \label{fig:MTO_select}
\end{figure*}

\begin{figure*}[h]
    \centering
    \caption{MTO, reject-only, with dummy stimuli}
    \includegraphics[height=0.4\textheight]{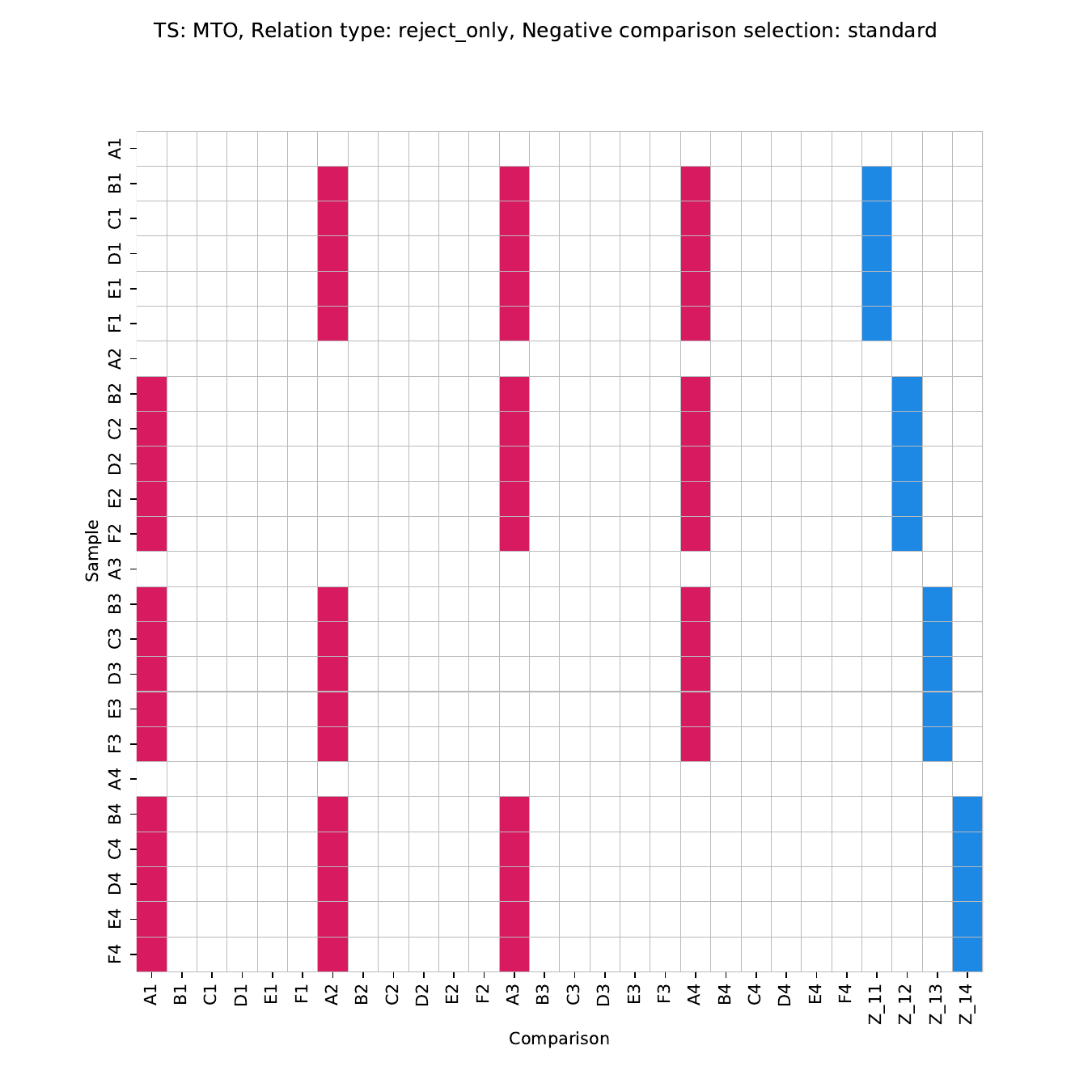}
    \label{fig:MTO_reject}
\end{figure*}

\begin{figure*}[h]
    \centering
    \caption{MTO, biased}
    \includegraphics[height=0.4\textheight]{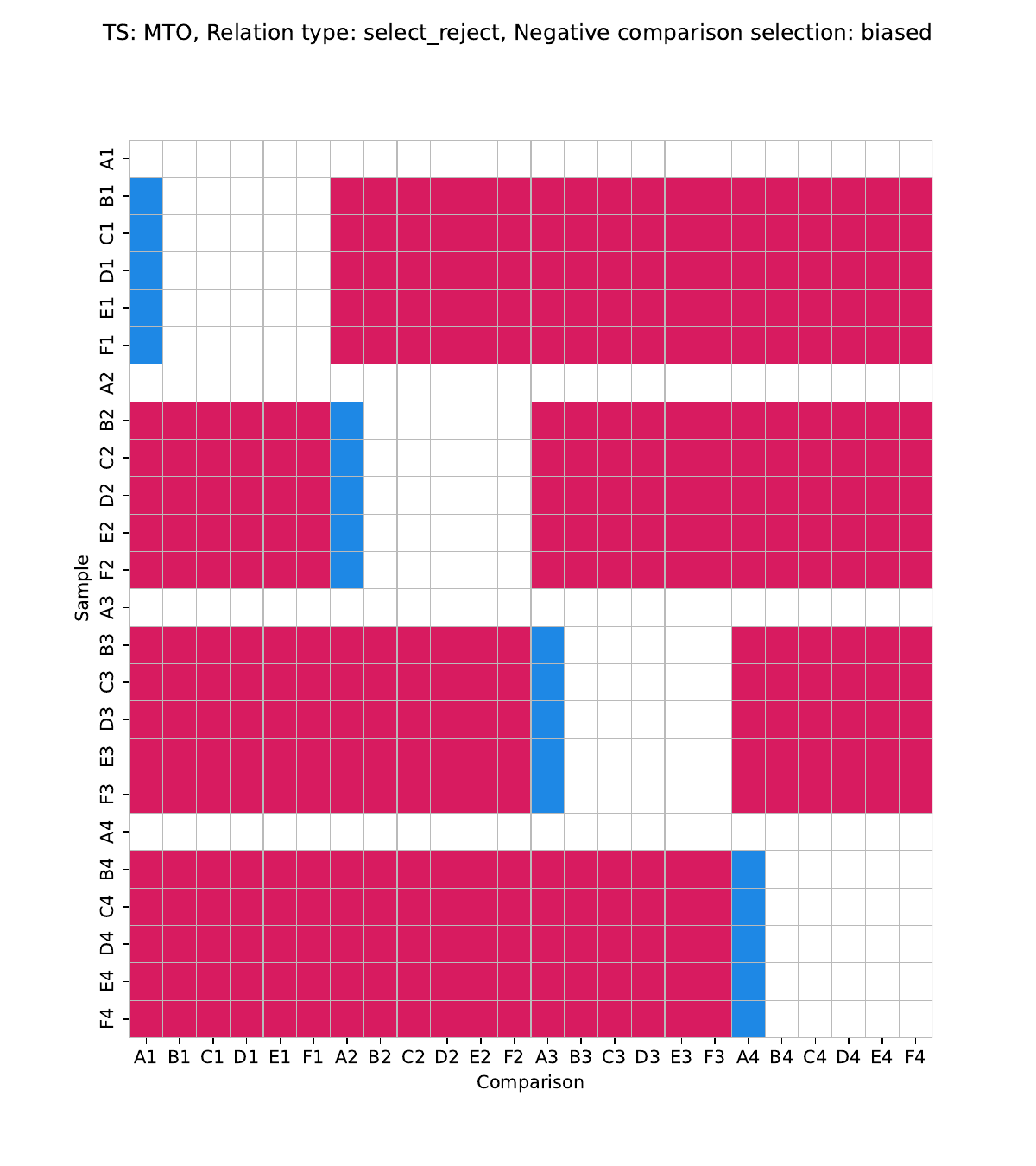}
    \label{fig:MTO_biased}
\end{figure*}

\begin{figure*}[h]
    \centering
    \caption{MTO, select-only, biased, with dummy stimuli}
    \includegraphics[height=0.4\textheight]{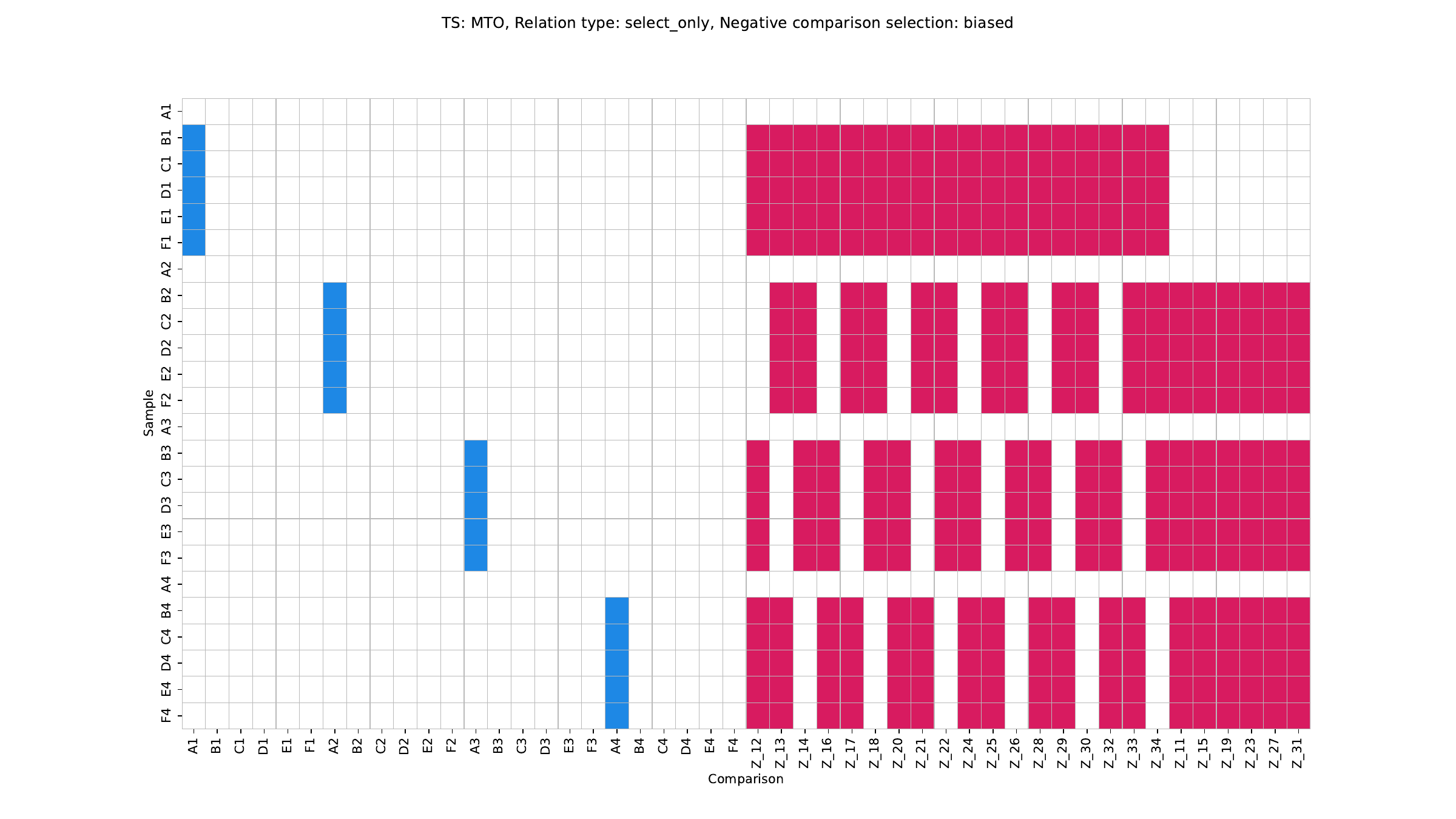}
    \label{fig:MTO_select_biased}
\end{figure*}

\begin{figure*}[h]
    \centering
    \caption{MTO, reject-only, biased, with dummy stimuli}
    \includegraphics[height=0.4\textheight]{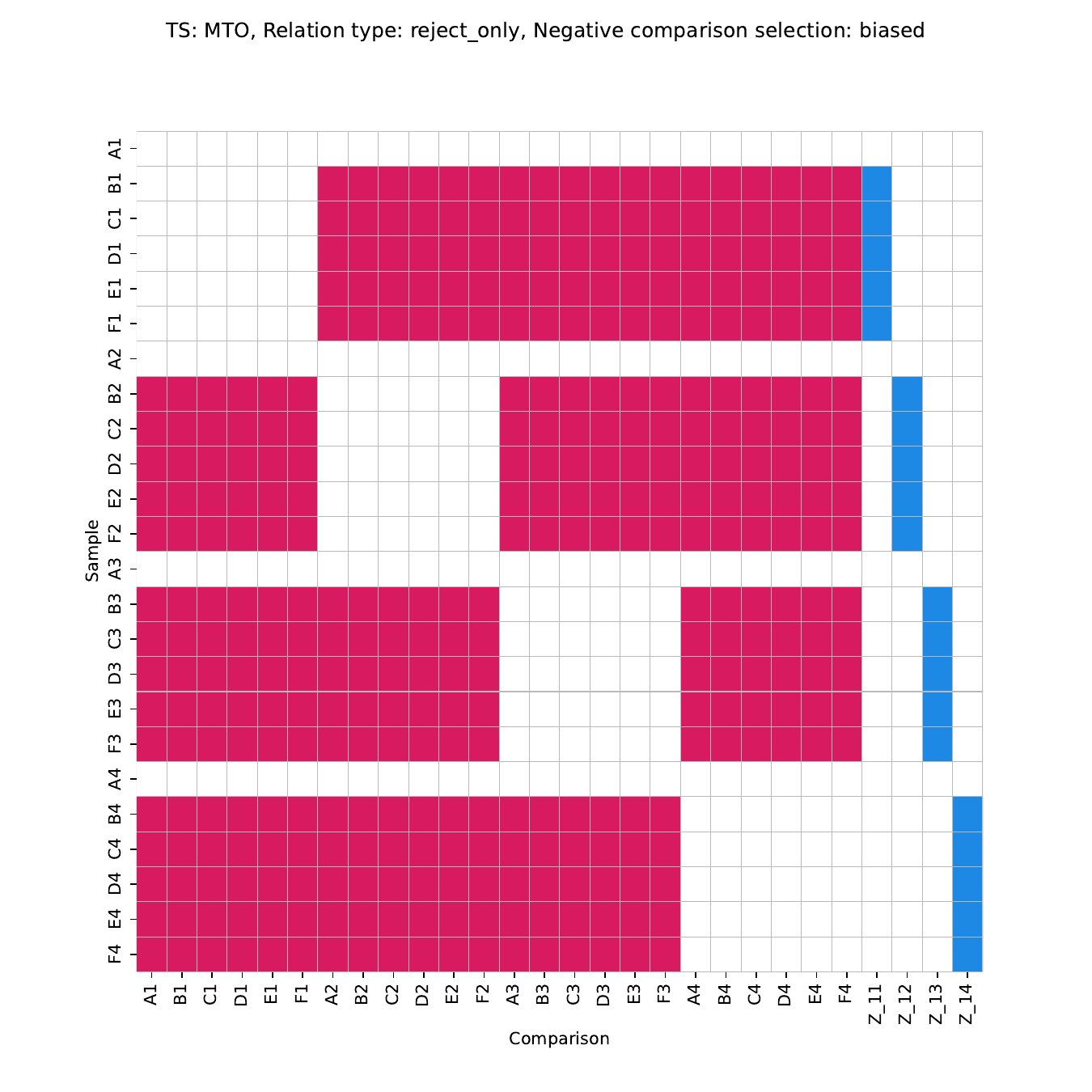}
    \label{fig:MTO_reject_biased}
\end{figure*}

\end{document}